\pdfoutput=1

\documentclass[11pt]{article}

\usepackage{acl}

\usepackage{times}
\usepackage{latexsym}

\usepackage[T1]{fontenc}

\usepackage[utf8]{inputenc}

\usepackage{microtype}

\usepackage{xcolor}
\usepackage{url}
\usepackage{amsmath}
\usepackage{amssymb}
\usepackage{mathtools}
\usepackage{graphicx}
\usepackage{booktabs}
\usepackage{multirow}
\usepackage{makecell}
\usepackage{xcolor}
\usepackage{bigdelim}
\usepackage{paralist}
\usepackage{relsize}
\usepackage{hyperref}
\usepackage{tikz,tikz-qtree}
\usepackage{pifont}
\usepackage{CJKutf8}
\usepackage{colortbl}

\definecolor{darkgreen}{RGB}{0,130,0}
\definecolor{cadmiumgreen}{rgb}{0.0, 0.42, 0.24}
\definecolor{alizarin}{rgb}{0.82, 0.1, 0.26}
\definecolor{pearDark}{HTML}{2980B9}
\DeclareMathOperator*{\argmax}{arg\,max}

\newenvironment{Japanese}{%
  \CJKfamily{min}%
  \CJKtilde
  }

\newcommand{\cmark}{\text{\ding{51}}}
\newcommand{\xmark}{\text{\ding{55}}}
\newcommand{\ptb}{\textsc{PTB}}
\newcommand{\ctb}{\textsc{CTB}}
\newcommand{\ktb}{\textsc{KTB}}
\newcommand{\roberta}{RoBERTa}

\newcommand{\para}[1]{\vskip 1mm\noindent\textbf{#1}~~}
\newcommand{\emldisplay}[2]{\texttt{\href{mailto:#1}{#2}}}
\newcommand{\eml}[1]{\emldisplay{#1}{#1}}

\newcommand{\ignore}[1]{}

\title{Co-training an Unsupervised Constituency Parser \\ with Weak Supervision}

\author{Nickil Maveli \and Shay B. Cohen \\
Institute for Language, Cognition and Computation \\
School of Informatics, University of Edinburgh \\
10 Crichton Street, Edinburgh, EH8 9AB \\
\eml{n.maveli@sms.ed.ac.uk},
\eml{scohen@inf.ed.ac.uk}}

\begin{document}
\maketitle
\begin{abstract}
We introduce a method for unsupervised parsing that relies on bootstrapping classifiers to identify if a node dominates a specific span in a sentence. There are two types of classifiers, an inside classifier that acts on a span, and an outside classifier that acts on everything outside of a given span. Through self-training and co-training with the two classifiers, we show that the interplay between them helps improve the accuracy of both, and as a result, effectively parse. A seed bootstrapping technique prepares the data to train these classifiers. Our analyses further validate that such an approach in conjunction with weak supervision using prior branching knowledge of a known language (left/right-branching) and minimal heuristics injects strong inductive bias into the parser, achieving 63.1 F$_1$ on the English (PTB) test set. In addition, we show the effectiveness of our architecture by evaluating on treebanks for Chinese (CTB) and Japanese (KTB) and achieve new state-of-the-art results.\footnote{Our code and pre-trained models are available at\\
\url{https://github.com/Nickil21/weakly-supervised-parsing}.}
\end{abstract}

\section{Introduction}
\label{sec:intro}

Pre-trained language models (PLMs) have become a standard tool in the Natural Language Processing (NLP) toolkit, offering the benefits of learning from large amounts of unlabeled data while providing modular function in many NLP tasks that require supervision. Recent work has shown that PLMs capture different types of linguistic regularities and information, for instance, the lower layers capture phrase-level information which becomes less prominent in the upper layers \citep{jawahar-etal-2019-bert}, span representations constructed from these models can encode rich syntactic phenomena, like the ability to track subject-verb agreement \citep{Goldberg2019AssessingBS}, dependency trees can be embedded within the geometry of BERT's hidden states \citep{hewitt-manning-2019-structural}, and most relevantly to this paper, syntactic information via self-attention mechanisms \citep{wang-etal-2019-tree, Kim2020Are}. 

\begin{figure}[t]
    \centering
     \includegraphics[width=0.7\linewidth]{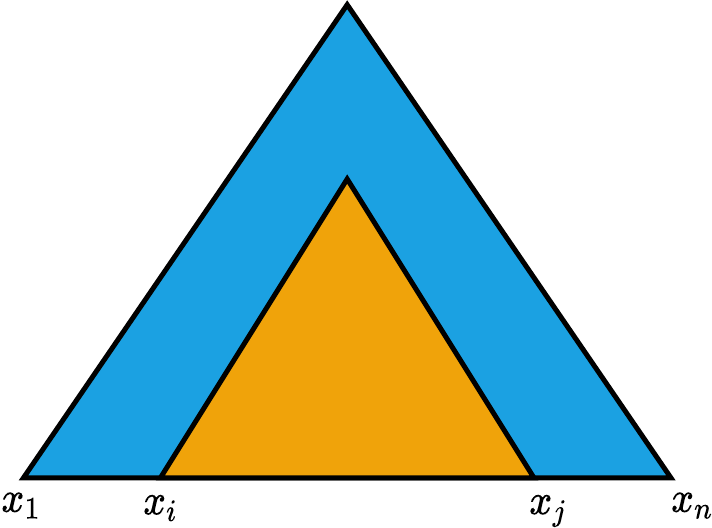}
    \caption{A depiction of a syntax tree, with the inside string as depicted by the sequence $x_i \cdots x_j$ and the outside string as depicted by the sequence $(x_1 \cdots x_{i-1}, x_{j+1} \cdots x_n)$ that provides external context for the inside representations.}
    \label{fig:io-tree}
\end{figure}

We offer another perspective on the way PLMs represent syntactic information. We demonstrate the usability of PLMs to capture syntactic information by developing an unsupervised parsing model that makes heavy use of PLMs. The learning algorithm is light in the injection of hard bias to parse text, emphasizing the role of PLMs in capturing syntactic information.

Our approach to unsupervised parsing is inspired by recent work in the area of spectral learning for parsing \citep{JMLR:v15:cohen14a, cohen-etal-2013-experiments} and unsupervised estimation of probabilistic context-free grammars (PCFGs; \citealp{clark-fijalkow-2020-consistent}). At its core, our learning algorithm views the presence or absence of a node dominating a substring in the final parse tree as a latent variable, where patterns of co-occurrence of the string that the node dominates (the ``inside'' string) and the rest of the sentence (the ``outside'' string) dictate whether the node is present or not. With spectral learning for latent-variable PCFGs (L-PCFGs; \citealp{JMLR:v15:cohen14a,cohen-collins-2014-provably}) the notion of inside \emph{trees} versus outside \emph{trees} is important, but in our case, given that the trees are not present during learning, we have to further specialize it to extract information only from the strings.

Consider the diagram of a syntax tree in Figure~\ref{fig:io-tree}, decomposed into two parts. Following the main notion in spectral learning, each of these parts (the orange part and the blue part) is a ``view'' of the whole tree that provides information on the identity of the node that spans the words $x_i \cdots x_j$. In the case of the tree being unobserved during training, we have to rely only on the substrings that are spanned by the blue part or the orange part, to hypothesize whether indeed a node exists there.

To represent the inside and outside views, we make use of PLMs. We encode these substrings, and then bootstrap a classifier that determines whether a given span is a constituent or not. The bootstrapping process alternates between the two views, and at each point adds predictions on the training set that it is confident about to train a new classifier. This can be thought of as a form of co-training \citep{yarowsky-1995-unsupervised, 10.1145/279943.279962}, a training technique that relies on multiple views of training instances.
We formulate the task of identifying constituents and distituents (referring to spans that are not constituents) in a sentence as a binary classification task by devising a strategy to convert the unlabeled data into a classification task. Firstly, we build a sequence classification model by fine-tuning a Transformer-based PLM on the unlabeled training sentences to distinguish between the true and false inside strings of constituents. Secondly, we use the highly-confident inside strings to produce the outside strings. Additionally, through the use of semi-supervised learning techniques, we jointly use both the inside and outside passes to enrich the model's ability to determine the breakpoints in a sentence. Our final model achieves 63.1 sentence F$_1$ averaged over multiple runs with random seed on the Penn Treebank test set. 
We also report strong results for the Japanese and Chinese treebanks.
\section{Problem Formulation and Inference}
\label{sec:Parsing-Algorithm}

We give a treatment to the problem of unsupervised constituency parsing. In that setup, the training algorithm is given an unlabeled corpus (set of sentences) and its goal is to learn a function mapping a sentence $x$ to an unlabeled phrase-structure tree $y$ that indicates the constituents in $x$. In previous work with models such as the Constituent-Context Model (CCM; \citealt{klein-manning-2002-generative}), the Dependency Model with Valence (DMV; \citealt{Klein2005NaturalLG}), and Unsupervised Maximum
Likelihood estimator for Data-Oriented Parsing (UML-DOP; \citealt{bod-2006-subtrees}), the parts of speech (POS) of the words in $x$ are also given as input both during inference and during training, but we do not make use of such POS tags.

\paragraph{Inference}
While our learning algorithm is grammarless, for inference we make use of a dynamic programming algorithm, akin to CYK, to predict the parse tree. Inference assumes that each possible span in the tree was scored with a score function $s(i,j)$ where $i$ and $j$ are endpoints in the sentence. The score function is learned through our algorithm. We then proceed by finding the tree $t^{\ast}$ such that:
\begin{align*}
t^{\ast} = \argmax_{t \in \mathcal{T}} \sum_{(i,j) \in t} s(i,j),
\end{align*}
\noindent where $\mathcal{T}$ is the set of possible binary trees over the sentence and $(i,j) \in t$, with a slight abuse of notation, denotes that the span $(i,j)$ appears in $t$.

When $s(i,j)$ is the probability of a span $(i,j$) being in the correct tree, this formulation gives the tree with the highest expected number of correct constituents~\citep{goodman-1996-parsing}. This formulation has been used recently by several unsupervised constituency parsing algorithms \citep{kim-etal-2019-unsupervised, kim-etal-2019-compound, cao-etal-2020-unsupervised, li-etal-2020-heads}.

\section{Training Algorithm}
\label{sec:training}

{\bf

\ignore{
\begin{itemize}
    \item section 3.1 is very obfuscated. what are you doing there exactly? building an inside classifier - are you using self-training there? there is no reference to any bootstrapping seeds, or reference to section 4.2.1.
    \item in section 3.2 - is this co-training yet? if you are using both the inside and the outside, how is it different than 3.3?
    \item I am completely lost on how you differentiate between the self-training algorithm and the co-training algorithm.
    \item In the experiment results, why are there no results for *both* the self-training *and* the co-training algorithm? (under Ours)? You need to have a more thorough investigation of these results.
    \item in the algorithm in Figure 2 - where did you get f in the beginning? you need to initialize it somehow - refer to the bootstrap or take it as input?
    \item In figure 2, why do you have an inside loop of prediction ('Predict on unlabeled...') of k steps, *and* an outer loop at the top of K iterations? ('Loop for K iteration...')
    \item You never mentioned where you get the Labeled inside set I, do you?
    \item What is $f(O)$?? Notationally this means applying $f$ on the set $O$ and returning the functions image on that set.
    \item why do you return just the outside classifier and not the inside classifier too in Figure 2?
\end{itemize}

the main issue is that after reading section 3.1-3.3 I am *more* confused about what you are doing! this needs to be thought out a bit more.
}
}

At the core of our approach lies the notion of \emph{inside} and \emph{outside} strings. For a given sentence $x = x_1 \cdots x_n$ and a span $(i,j)$, the inside string of span $(i,j)$ is the sequence $x_i \cdots x_j$ while the outside string is the pair $(x_1 \cdots x_{i-1}, x_{j+1} \cdots x_n)$. We denote by ${\boldsymbol{h}_{\text{in}}(i,j)}$ representations for inside strings and ${\boldsymbol{h}_{\text{out}}(i,j)}$ representations for outside strings. Both are vectors derived from a PLM (RoBERTa; \citealt{DBLP:journals/corr/abs-1907-11692}, as we see later).



These two types of strings provide two views of a given possible splitting point in the syntax tree. We offer three ways, with increasing complexity, to bootstrap a score function that helps identify whether a node should dominate a given span.
The main idea behind this bootstrapping is to start with a small seed set of training examples $(x, i,j, b)$ where $(i,j)$ is a span in a sentence $x$ and $b \in \{ 0, 1 \}$, depending on whether the span $(i,j)$ is dominated by a node in the syntactic tree or not.
Bootstrapping the seed set is dependent only on either the inside string or the outside string, and the corresponding classifier built from this bootstrapped seed set returns a probability $p(b \mid x, i, j)$. Once a classifier is learned using the bootstrapping seed set, the classifier is applied on the training set, and the seed set is added to more examples where the classifier is confident of the label $b$. This is also known as self-training \citep{mcclosky-etal-2006-effective, mcclosky-etal-2008-self}.

\ignore{
\begin{figure}[t]
\begin{footnotesize}
\framebox{\parbox{\columnwidth}{

{\bf Inputs:} Input strings $x^{(1)},\ldots,x^{(N)}$, an inside seed function $\mathrm{iseed}(x,i,j) \in \{ 0, 1, \bot \}$ and an outside seed function $\mathrm{oseed}(x,i,j) \in \{ 0, 1, \bot \}$.

{\bf Algorithm:} (calculate all inside and outside strings)

\begin{itemize}

\item For all $i \in [N]$ such that $a_i \in $,

\end{itemize}

{\bf Return:}
${f^1} c^1_{a_1}$
}}
\end{footnotesize}
\caption{\small The tensor form for calculation of $p(r_1 \ldots
  r_N)$.}
\label{fig:dpobservable}
\vspace{-2ex}
\end{figure}
}

In the next three sections, we present three learning algorithms of increasing complexity in their use of inside and outside strings.

\subsection{Modeling Using Inside Strings}
\label{ssec:inside-algorithm}
The inside model $m_{\text{in}}$ which is modeled at a sentence level, computes an inside score $s_{\text{in}}{(i,j)}$ from the inside vector representation ${\boldsymbol{h}_{\text{in}}(i,j)}$ of each span in the unlabeled input training sentence $\mathcal{\mathbf{U}}$. To compute ${\boldsymbol{h}_{\text{in}}(i,j)}$, we fine-tune the sequence classification model that encodes a fixed-vector representation for each token in the dataset.\ignore{$\leftarrow$ I think the reviewers intention here was that it needs to be clear what you use - in this case ROBERTa, so maybe just briefly mention ROBERTa in a footnote and say that it will become more clear later. finally, that sentence "To compute..." is not clear to me. Can you explain to me in an email what you mean?} This captures the phrase information of the inner content in the span. In order to prepare the features for the inside model, we make use of a seed bootstrapping technique (Section~\ref{ssec:seed-bootstrap}). Once we build the inside model $m_{\text{in}}$, we get the most confidently-classified inside strings from $\mathcal{\mathbf{U}}$ based on a set threshold $\tau=(\tau_{\text{min}}, \tau_{\text{max}})$. Here, $\tau_{\text{min}}$ and $\tau_{\text{max}}$, form the confidence bounds to select distituents and constituents respectively. We select a random sample of $c$ constituents and $d$ distituents with appropriate labels from these most confident inside strings comprising the labeled inside set $\mathcal{\mathbf{I}}$.


\begin{figure}[t]
\centering
\begin{small}
\framebox{\parbox{0.95\linewidth}{
{\bf Inputs:} 
$\mathcal{\mathbf{I}}$ represents the labeled inside set;
$\mathcal{\mathbf{U}}$ is a set of unlabeled training sentences;\\

{\bf Algorithm:}

\begin{compactitem} 
\item Loop for $K$ iterations:
\begin{enumerate}
\item Learn the inside classifier $m_{\text{in}}$ based on ${\boldsymbol{h}_{\text{in}}(i,j)}$ derived from $\mathcal{\mathbf{I}}$
\item Use $m_{\text{in}}$ to label $\mathcal{\mathbf{U}}$ to get the predicted inside strings $\hat{y}_{\text{in}}$
\item If $\hat{y}_{\text{in}}$ $>$ $\tau_{\text{max}}$, extract $c$ constituents randomly and add it to the set of pseudo-constituents $\mathcal{X}_c$
\item If $\hat{y}_{\text{in}}$ $<$ $\tau_{\text{min}}$, extract $d$ distituents randomly and add it to the set of pseudo-distituents $\mathcal{X}_d$
\item $\mathcal{\mathbf{I}} = \mathcal{X}_c \cup \mathcal{X}_d$
\end{enumerate}
\item Get outside strings for each $\mathcal{\mathbf{I}}$; Assign to the set of labeled output sentences $\mathcal{\mathbf{O}}$
\item Learn outside model $m_{\text{out}}$ based on ${\boldsymbol{h}_{\text{out}}(i,j)}$ derived from $\mathcal{\mathbf{O}}$
\end{compactitem}
\bigskip

{\bf Output:}
inside model $m_{\text{in}}$, outside model $m_{\text{out}}$}
}
\end{small}
\caption[Our self-training algorithm]{Our self-training algorithm.}
\label{fig:our-self-training-algo}
\end{figure}

\subsection{Modeling Using Inside and Outside Strings}
\label{ssec:inside-outside-algorithm}
 To perform the iterative self-training procedure, we follow the steps as detailed in Figure~\ref{fig:our-self-training-algo}. While building the outside model, we extract the tokens at the span boundaries of the pair of outside strings, which is of the form consisting of the triple ($x_{i-1}$, \texttt{[MASK]}, $x_{j+1}$). The outside model computes an outside score $s_{\text{out}}{(i,j)}$ from the outside vector representation ${\boldsymbol{h}_{\text{out}}(i,j)}$ of each span, which models the contextual information of the span. To compute ${\boldsymbol{h}_{\text{out}}(i,j)}$, we extract the triple for every span $(i,j)$ in the dataset and fine-tune another sequence classification model that encodes a fixed-vector representation for each triple.


\subsection{An Iterative Co-training Algorithm}
\label{ssec:co-training-algorithm}
 Co-training \citep{10.1145/279943.279962} is a classic multi-view training method, which trains a classifier by exploiting two (or more) views of the training instances. Our final learning algorithm is indeed inspired by it, where we consider the inside and the outside strings to be the two views. Once we have the inside $m_{\text{in}}$ and the outside classifiers $m_{\text{out}}$ that are trained on their respective conditionally independent inside ${\boldsymbol{h}_{\text{in}}(i,j)}$ and outside ${\boldsymbol{h}_{\text{out}}(i,j)}$ feature sets, we can make use of an iterative approach. At each iteration, only the inside strings $\hat{\mathcal{\mathbf{I}}}$ that are confident to be likely the insides of constituents and distituents according to the outside model are moved to the labeled training set of the inside model $\mathcal{\mathbf{I}}$. Thus, the outside model (teacher) provides the labels to the inside strings on which the inside model (student) is uncertain. Similarly, only the outside strings $\hat{\mathcal{\mathbf{O}}}$ that are confident to be the likely outsides of constituents and distituents according to the inside model are moved to the labeled training set of the outside model $\mathcal{\mathbf{O}}$. Thus, the inside model provides the labels to the outside strings on which the outside model is uncertain. We describe the steps in Figure~\ref{fig:our-co-training-algo}.
Finally, we combine the scores obtained by the inside and the outside model to get the score ${s(i,j)}$ for each span:
\begin{align*}
    s(i,j) = s_{\text{in}}(i,j) \cdot s_{\text{out}}(i,j).
\end{align*}

\begin{figure}[t]
\centering
\begin{small}
\framebox{\parbox{\linewidth}{
{\bf Inputs:}
$\mathcal{\mathbf{I}}$ is the set of labeled inside sentences;
$\mathcal{\mathbf{O}}$ is the set of labeled outside sentences;
$\mathcal{\mathbf{U}}$ is a set of unlabeled sentences.
\newline

{\bf Algorithm:} Loop for $K$ iterations:
\begin{compactitem}
\item Choose $c$ pseudo-constituents and $d$ pseudo-distituents from the most confidently predicted outside strings $\hat{y}_{\text{out}}$ from $\mathcal{\mathbf{U}}$ based on $\tau$
\item Extract the inside strings $\hat{\mathcal{\mathbf{I}}}$ corresponding to the $c$ pseudo-constituents and $d$ pseudo-distituents of outside
\item  $\mathcal{\mathbf{I}} = \mathcal{\mathbf{I}} \cup \hat{\mathcal{\mathbf{I}}}$
\item Train the inside model $m_{\text{in}}$ based on ${\boldsymbol{h}_{\text{in}}(i,j)}$ derived from $\mathcal{\mathbf{I}}$
\item Choose $c$ pseudo-constituents and $d$ pseudo-distituents from the most confidently predicted inside strings $\hat{y}_{\text{in}}$ from $\mathcal{\mathbf{U}}$ based on $\tau$
\item Extract the outside strings $\hat{\mathcal{\mathbf{O}}}$ corresponding to the $c$ pseudo-constituents and $d$ pseudo-distituents of inside
\item $\mathcal{\mathbf{O}} = {\mathcal{\mathbf{O}}} \cup \hat{\mathcal{\mathbf{O}}}$
\item Train the outside model $m_{\text{out}}$ based on ${\boldsymbol{h}_{\text{out}}(i,j)}$ derived from $\mathcal{\mathbf{O}}$
\end{compactitem}
\bigskip

{\bf Output:}
Two models $m_{\text{in}}$, $m_{\text{out}}$, that predict the inside and outside scores for unlabeled sentences. We combine these predictions by multiplying together and optionally re-normalizing their class probability scores.}}
\end{small}
\caption[Our co-training algorithm]{Our co-training algorithm.}
\label{fig:our-co-training-algo}
\end{figure}

Co-training requires the two views to be independent of each other conditioned on the label of the training instance. This is the type of assumption that, for example, PCFGs satisfy, when breaking a tree into an outside and inside tree: the two trees are conditionally independent given the nonterminal that connects them. In our case, we satisfy this assumption by creating inside and outside string representations separately, as we see later in Section~\ref{sec:experiments}.

\begin{figure*}[t]
    \centering
     \includegraphics[width=1.\linewidth]{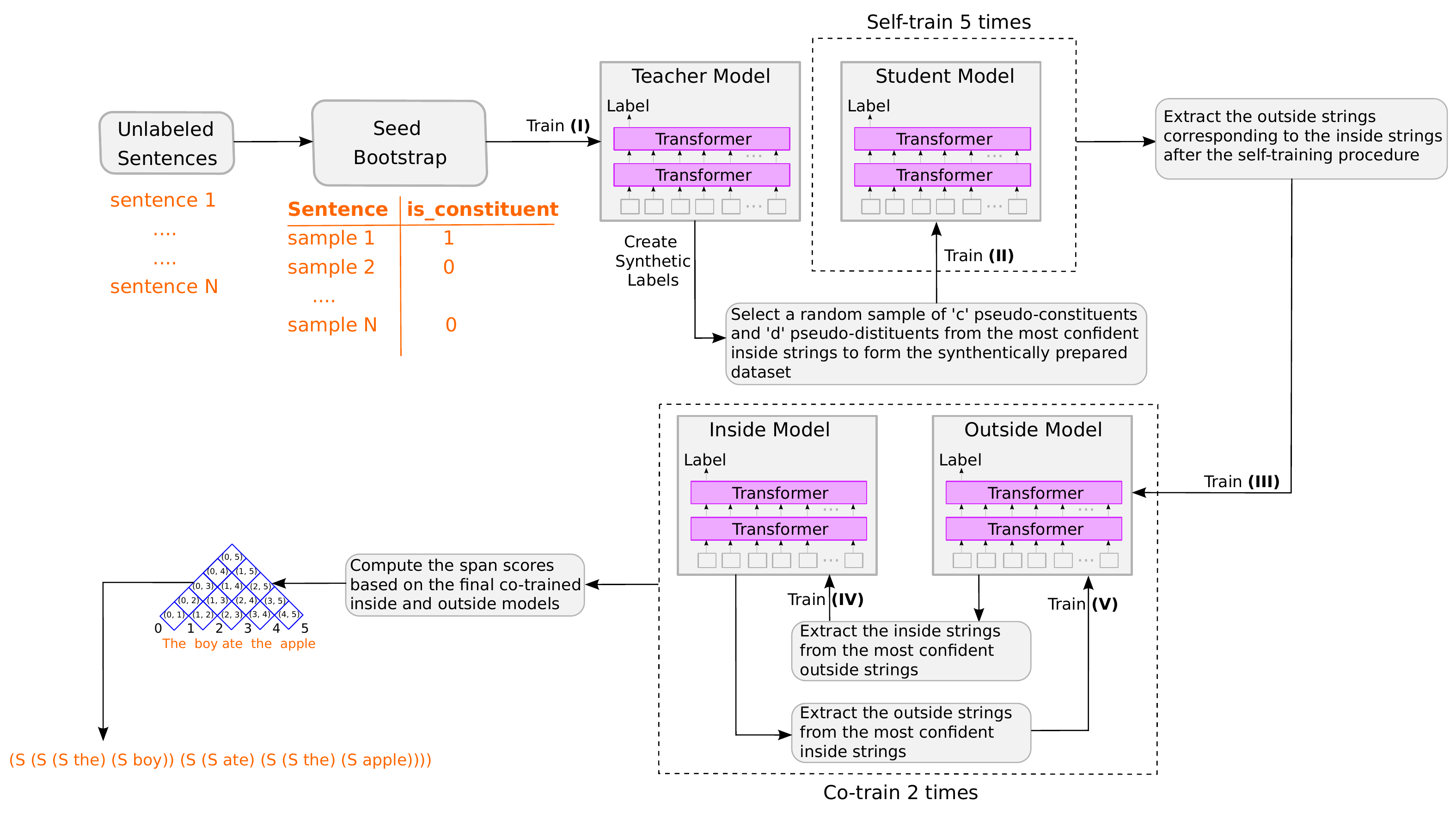}
    \caption[Block diagram detailing our approach]{Block diagram detailing our approach. We perform the self-training procedure for five iterations which follow multiple steps; \textbf{(I)}: Fine-tune a RoBERTa\textsubscript{\textsc{Base}} model (teacher) on a downstream task using a cross-entropy loss after seed bootstrapping; \textbf{(II)}: Synthetically annotate this data using the teacher model and select top K samples corresponding to each class to form the final synthetic dataset; We fine-tune a RoBERTa\textsubscript{\textsc{Base}} model (student) on this dataset using hard labels and retrieve the outside strings from the most confident insides; \textbf{(III)}: Train the outside classifier on these outside strings;  We perform the co-training procedure for two iterations which follow a two-fold optimizing step; \textbf{(IV)}: Retrieve the inside strings from the most confident outsides and train the inside classifier; \textbf{(V)}: Retrieve the outside strings from the most confident insides and train the outside classifier.}
    \label{fig:block-diagram}
\end{figure*}

Figure~\ref{fig:block-diagram} illustrates the underlying pipeline of our weakly supervised parsing framework in an end-to-end fashion.

\section{Experimental Setup}
\label{sec:experiments}

In this section, we describe our experimental setup: the data we use, the exact details of the experimental use of our approach to unsupervised parsing, and our evaluation methodology.

\subsection{Data}
\label{ssec:data}
We evaluate our methodology on the Penn Treebank (\ptb{}; \citealt{marcus-etal-1993-building}) with the standard splits (2-21 for training, 22 for validation, 23 for test). For preprocessing, we keep all punctuation and remove any trailing punctuation. To maintain the unsupervised nature of our experiments, we avoid the common practice of using gold parses of the validation set for either early stopping \citep{shen2018neural, shen2018ordered, drozdov-etal-2019-unsupervised} or hyperparameter tuning \citep{kim-etal-2019-compound}. Additionally, we experiment on Chinese with version 5.1 of the Chinese Penn Treebank (\ctb{}; \citealt{DBLP:journals/nle/XueXCP05}) with the same splits as in \citet{chen-manning-2014-fast}, and the Japanese Keyaki Treebank (\ktb{}; \citealt{butlerkeyaki}). For \ktb, we shuffle the corpus and use 80$\%$ of the sentences for training, 10$\%$ for validation, and 10$\%$ for testing.

\subsection{Multi-view Learning}
\label{ssec:multi-view-learning}
In this section, we devise the task of identifying constituents in a sentence by training two models with different views of the data. Ideally, these views complement each other and help each model improve the performance of the other.

\subsubsection{Seed Bootstrapping}
\label{ssec:seed-bootstrap}

We treat identifying constituents from unlabeled sentences as a sequence classification task. To generate the \emph{constituent} class, we take the complete sentence \begin{small}\texttt{(start:end)}\end{small}, as a sentence in itself is a constituent, and also the largest among all of its other constituents. To generate the \emph{distituent} class, we take \begin{small}\texttt{(start:end-1)}\end{small}, $\cdots$, \begin{small}\texttt{(start:end-6)}\end{small} slices, where \begin{small}{\texttt{start}}\end{small} and \begin{small}{\texttt{end}}\end{small} denote the 0\textsuperscript{th} and N\textsuperscript{th} position (sentence length) respectively. We select the distituents in this manner because the longer the sentence, there would be a significantly unlikely chance that the span of the constituents extends till the very end of the sentence.
Additionally, we make use of casing-specific information by adding contiguous title-case words while allowing only the apostrophe mark. Since all of the sentences for the constituent class start with capital letters, we identify the most common first word and generate lower-case equivalents of contiguous title-case words, which starts with it to account for bias due to the casing of spans. While we do use a fixed template to perform the seed bootstrapping process, this is part of the inductive bias of the algorithm, and is relatively easy to acquire. 
In our analysis, we assume the language is already known before and thereby its \emph{structure} (left/right-branching), a form of weak supervision.

For \ctb{}, we follow the exact same process as \ptb{} for preparing the input data for the first-level sequence classifier, but we do not rely on case-specific information and perform no post-processing. Meanwhile, since \ktb{} is a treebank of a strongly left-branching language, we design our modeling approach slightly differently compared to before, although along the same style. To prepare the data for the sequence classifier, we choose the slice \begin{small}\texttt{(start:end)}\end{small} in the sentence to label the \emph{constituent} class, whereas, \begin{small}{\texttt{(start+1:end)}}\end{small}, $\cdots$, \begin{small}{\texttt{(start+4:end)}}\end{small} slices are chosen to label the \emph{distituent} class. We also split the sentences on ``\texttt{*}'' mark and treat the resulting fragmented parts as constituents too. Our training does not depend on the development set with the gold-standard annotated trees since we base the necessary string slicing decision on the feedback from the validation split after the bootstrapping procedure in an iterative fashion (increment/decrement the value of slice counter by 1) until we see a degradation in performance (measured using F$_1$ score) on the synthetic set of seed constituents and distituents.

\subsubsection{Inside Model}
\label{ssec:inside-model-prepare}

We fine-tune the \roberta{} model with a sequence classification layer on top using a cross-entropy loss (see Section~\ref{appx:training-details} in Appendix for training and hyperparameter details). As we supply input data, the entire pre-trained \roberta{}\textsubscript{\textsc{base}} model and the additional untrained classification layer is trained on our specific downstream task. To compute $\boldsymbol{h}_{\text{in}}{(i, j)}$, we run the \roberta{}\textsubscript{\textsc{base}} model and retrieve the \texttt{[CLS]} token representation for the span enclosed between the $i^{\text{th}}$ and the $j^{\text{th}}$ element. The inside model is evaluated on MCC (Matthews Correlation Coefficient) as well as F$_1$ because the classes are imbalanced. After fine-tuning, our best inside model achieves 0.62 MCC and 0.91 F$_1$ on the internal validation set. Finally, we fine-tune the inside model on the unlabeled training sentences that generates an inside score $s_{\text{in}}{(i,j)}$ for every span. Since our major focus was on \ptb{}, we have listed a few \emph{heuristics} that inject further bias into the algorithm acting as the another form of weak supervision. Moreover, incorporating such rules was not necessary for \ctb{} and \ktb{} as our models showed superior performance without them.

Once we compute the inside score, $s_{\text{in}}{(i,j)}$, we use the following refinement strategies to prune out false constituents:
We delete any constituent if it starts or ends with the most common word succeeding the comma punctuation.
Additionally, we take the most common starting word and check if its accompanying word does not belong to the NLTK stop words list. We assign the scores of these corresponding spans in the CYK chart cell to the maximum value. Intuitively, from the linguistic definition of constituents, we refrain from bracketing if we identify a contiguous group of rare titlecase or uppercase words (tokens not in the top 100 most frequent list in the \ptb{} training sentences).
These heuristics only contribute to a certain extent in making the parser strong, and should be considered as a standard post-processing step. Overall, we observe 3.8 F$_1$ improvements in the case of the inside model. We further note that the contribution due to additional heuristics is much less than the combined self-training and co-training gains since their effect becomes insignificant after multiple iterations of the self-training process due to the predictions approximately following the template rules.
As described in Figure~\ref{fig:our-self-training-algo}, we perform self-training on the inside model for five iterations.\footnote{We only use the top 5K inside strings for self-training to cover maximum possible iterations as it is representative of the whole training set in terms of the average sentence length and punctuation marks.}

\subsubsection{Outside Model}
\label{ssec:outside-model-prepare}
We extract the outside strings of spans having the inside score satisfying a pre-determined cutoff value.  The Constituent-Context Model~\citep{klein-manning-2002-generative} use a smoothing ratio of 1:5 (constituents to distituents) for the WSJ-10 section to take into account the skewness of random spans more likely to represent distituents. In the same vein, the values of lower and upper bounds of the threshold are chosen to ensure the distribution of class labels is about 1:10 (with the distituent class being the majority) which is a crude estimate considering much larger sentence lengths in the WSJ-Full section.
Moreover, from a linguistic standpoint, we can be certain that the distituents must necessarily outnumber the constituents. For the self-training experiments, we set the thresholds, $\tau_{\text{min}}$ as 0.0005 and $\tau_{\text{max}}$ as 0.995. We treat the outside strings satisfying the upper and lower bounds of the threshold as \emph{gold-standard outside} of constituents and distituents respectively. To compute $\boldsymbol{h}_{\text{out}}{(i, j)}$, we run the \roberta{}\textsubscript{\textsc{base}} model on left-outside, i.e., ${(i-1)}^{\text{th}}$ element and right-outside, i.e., ${(j+1)}^{\text{th}}$ element, along with a \texttt{[MASK]} placeholder token separating the two, and extract the \texttt{[CLS]} token representation. As done previously, we fine-tune the outside model on the unlabeled training sentences that generates an outside score $s_{\text{out}}{(i,j)}$ for every span.

\subsubsection{Jointly Learning with Inside and Outside Models}
\label{ssec:joint-inside-outside}
Once we have the outside model, we run it on the training sentences and choose the outside string that the classifier is highly confident about. We extract their inside strings again using the same bounds of the threshold as done previously and re-train the inside model on the \emph{old highly confident inside strings} along with the \emph{new inside strings obtained from the highly confident outside strings}. Similarly, the same technique can be applied to the outside model to augment its input data too. We repeat this process twice (Figure~\ref{fig:our-co-training-algo}).

\begin{table}[t]
    \centering
    \resizebox{1.\linewidth}{!}{%
\begin{tabular}{lcccc}
    \toprule
    \multirow{2}{*}{\textbf{Model}} & \multicolumn{2}{c}{\textbf{WSJ-Full}} & \multicolumn{2}{c}{\textbf{WSJ-10}} \\
    & Mean & Max & Mean & Max \\
    \midrule
    \multicolumn{3}{l}{\emph{Trivial Baselines:}}\\
    \midrule
    \multicolumn{1}{l}{Left Branching (LB)} & 8.7 & & 17.4 & \\       
    \multicolumn{1}{l}{Balanced} & 18.5 & \\
    \multicolumn{1}{l}{Right Branching (RB)} & 39.5 & & 58.5 & \\   
    \midrule
    \multicolumn{3}{l}{\emph{Unsupervised Parsing approaches:}}\\
    \midrule
    \multicolumn{1}{l}{PRPN\textsuperscript{\textdagger} \citep{shen2018neural}} & 37.4 & 38.1 & 58.4 & -- \\
    \multicolumn{1}{l}{URNNG\textsuperscript{$\star$} \citep{kim-etal-2019-unsupervised}} & -- & 45.4 & -- & -- \\
    \multicolumn{1}{l}{ON\textsuperscript{\textdagger} \citep{shen2018ordered}} & 47.7 & 49.4 & 63.9 & -- \\
    \multicolumn{1}{l}{Tree Transformer\textsuperscript{\textdagger$\star$} \citep{wang-etal-2019-tree}} & 50.5 & 52.0 & 66.2 & -- \\
    \multicolumn{1}{l}{Neural PCFG\textsuperscript{\textdagger} \citep{kim-etal-2019-compound}} & 50.8 & 52.6 & 64.6 & -- \\
    \multicolumn{1}{l}{DIORA\textsuperscript{$\star$} \citep{drozdov-etal-2019-unsupervised}} & -- & 58.9 & 60.5 & -- \\
    \multicolumn{1}{l}{Compound PCFG\textsuperscript{\textdagger} \citep{kim-etal-2019-compound}} & 55.2 & 60.1 & 70.5 & -- \\
    \multicolumn{1}{l}{S-DIORA\textsuperscript{\textdagger$\star$} \citep{drozdov-etal-2020-unsupervised}} & 57.6 & 64.0 & 71.8 & -- \\
    \multicolumn{1}{l}{Constituency Test\textsuperscript{$\star$} \citep{cao-etal-2020-unsupervised}} & 62.8 & 65.9 & 68.1 & -- \\
    \multicolumn{1}{l}{Ours\textsuperscript{$\star$} (using inside)} & \cellcolor{pearDark!20}55.9 & \cellcolor{pearDark!20}57.2 & \cellcolor{pearDark!20}66.2 & -- \\
    \multicolumn{1}{l}{Ours\textsuperscript{$\star$} (using inside w/ self-training)} & \cellcolor{pearDark!20}61.4 & \cellcolor{pearDark!20}64.2 & \cellcolor{pearDark!20}71.7 & --\\
    \multicolumn{1}{l}{Ours\textsuperscript{$\star$} (using inside and outside w/ co-training)} & \cellcolor{pearDark!20}\textbf{63.1} & \cellcolor{pearDark!20}66.8 & \cellcolor{pearDark!20}\textbf{74.2} & --\\
    \midrule
    \multicolumn{1}{l}{Oracle Binary Trees} & 84.3 & & 82.1 & \\
    \bottomrule
    \end{tabular}
    }
    \caption[Results on the PTB test set]{Unlabeled sentence-level F$_1$ on the full as well as sentences of length $\leq$ 10 of the PTB test set without punctuation or unary chains. We evaluate each model using the evaluation script provided by \citet{kim-etal-2019-compound} and take the baseline numbers of certain models from \citep{kim-etal-2019-compound, cao-etal-2020-unsupervised}.
    {\textdagger} denotes models trained without punctuation and $\star$ denotes models trained on additional data.}
    \label{tab:ptb-results}
\end{table}
\begin{table}[t]
   \centering
    \resizebox{1.\columnwidth}{!}{%
    \begin{tabular}{lcc}
    \toprule
    \multirow{2}{*}{\textbf{Model}} & \multicolumn{2}{c}{\textbf{CTB}} \\ & Mean & Max \\
    \midrule
    \multicolumn{3}{l}{\emph{Trivial Baselines:}}\\
    \midrule
    \multicolumn{1}{l}{Left Branching (LB)} & 9.7 & \\       
    \multicolumn{1}{l}{Random Trees} & 15.7 & 16.0 \\
    \multicolumn{1}{l}{Right Branching (RB)} & 20.0 & \\ 
    \midrule
    \multicolumn{3}{l}{\emph{Unsupervised Parsing approaches:}}\\
    \midrule
    \multicolumn{1}{l}{PRPN~\citep{shen2018neural}} & 30.4 & 31.5 \\
    \multicolumn{1}{l}{ON~\citep{shen2018ordered}} & 25.4 & 25.7 \\
    \multicolumn{1}{l}{Neural PCFG~\citep{kim-etal-2019-compound}} & 25.7 & 29.5 \\
    \multicolumn{1}{l}{Compound PCFG~\citep{kim-etal-2019-compound}} & 36.0 & 39.8 \\
    \multicolumn{1}{l}{Ours (using inside)} & \cellcolor{pearDark!20}37.8 & \cellcolor{pearDark!20}38.4 \\
    \multicolumn{1}{l}{Ours (using inside w/ self-training)} & \cellcolor{pearDark!20}40.6 & \cellcolor{pearDark!20}41.7 \\
    \multicolumn{1}{l}{Ours (using inside and outside w/ co-training)} & \cellcolor{pearDark!20}\textbf{41.8} & \cellcolor{pearDark!20}43.3  \\
    \midrule
    \multicolumn{1}{l}{Oracle Binary Trees} & 81.1 & \\
    \bottomrule
    \end{tabular}
    }
    \caption[Results on the CTB test set]{Unlabeled sentence-level F$_1$ on the CTB test set. We evaluate each model using the evaluation script provided by \citet{kim-etal-2019-compound} and take the baseline numbers also from \citet{kim-etal-2019-compound}.}
    \label{tab:ctb-results}
\end{table}

\subsection{Evaluation}
\label{ssec:evaluation}
We report the F$_1$ score with reference to gold trees in the \ptb{} test set (section 23). Following prior work \citep{kim-etal-2019-compound, shen2018neural, shen2018ordered, cao-etal-2020-unsupervised}, we remove punctuation and collapse unary chains before evaluation, and calculate F$_1$ ignoring trivial spans, i.e., single-word spans and whole-sentence spans, and we perform the averaging at sentence-level (macro average) rather than span-level (micro average), which means that we compute F$_1$ for each sentence and later average over all sentences. We also mention the oracle upper bound, which is the highest possible score with binarized trees since we compare them against non-binarized gold trees according to the convention, as most unsupervised parsing methods output fully binary trees. We additionally use the standard PARSEVAL metric computed by the \texttt{evalb} program.\footnote{\url{https://nlp.cs.nyu.edu/evalb}} Although \texttt{evalb} calculates the micro average F$_1$ score, it differs from our micro average metric in the sense that it counts the whole sentence spans, and calculates duplicated spans instead of removing them. Following the recommendations put forth by previous work that has done a comprehensive empirical evaluation on this topic~\citep{li-etal-2020-empirical}, we report results on both length $\leq$ 10 as well as all-length test data.

\section{Results and Discussion}
\label{sec:results}

Table~\ref{tab:ptb-results} shows the unlabeled F$_1$ scores for our model compared to existing unsupervised parsers on \ptb{}. The vanilla inside model is in itself competitive and is already in the range of previous best models like DIORA~\citep{drozdov-etal-2019-unsupervised}, Compound PCFG~\citep{kim-etal-2019-compound}.\footnote{We do not include the results of \citet{shi-etal-2021-learning} in our analysis because their boost in the performance is contingent on the nature of the supervision data (especially the QA-SRL dataset) rather than on the actual learning process itself. Furthermore, the authors mention that a vast amount of hyperlinks match syntactic constituents, hence restricting the scope for the actual algorithm to derive meaningful trees.} See Appendix~\ref{appx:unsup-labeled-parsing} to assess our model's performance on unsupervised labeled parsing.

We further evaluate how our method works for languages with different branching types -- Chinese (right-branching) and Japanese (left-branching).
\ignore{
The issue with multilingual \textsc{BERT} (\textsc{mBERT}; \citealp{devlin-etal-2019-bert}) as well as with \textsc{XLM-RoBERTa} \citep{conneau-etal-2020-unsupervised} is that those produce rather bad sentence representation out-of-the-box. Further, the vector spaces between languages are not aligned, i.e., the sentences with the same content in different languages would be mapped to different locations in the vector space. Given that we were more easily able to achieve good results with monolingual Transformer models, and knowing that it requires less memory (due to a smaller vocabulary), we decided to stick with it (see Section~\ref{appx:training-details} in Appendix for training details).
}
We use Transformer models for the representations of the spans for both Chinese and Japanese. See Section~\ref{appx:training-details} in the Appendix for training details. Tables~\ref{tab:ctb-results} and ~\ref{tab:ktb-results} shows the results for \ctb{} and \ktb{} respectively. Moreover, we do not include a few models chosen previously for \ptb{} during our analysis, as extending those models for \ctb{} or \ktb{} is non-trivial due to several reasons: such as lack of domain-related datasets (as {DIORA} uses SNLI and MultiNLI for training), and lack of linguistic knowledge expertise (not easily cross-lingual transferable notion for designing constituency tests).

Figure~\ref{fig:ptb-trees} in the Appendix shows step-wise qualitative analysis for a sample sentence taken from the \ptb{} training set. See Figures~\ref{fig:ctb-trees} and~\ref{fig:ktb-trees} in Appendix to see the visualization for an example tree at every stage of the pipeline for \ctb{} and \ktb{} respectively. As we can observe from all the example tree outputs, the parser using the inside and outside models after the co-training stage produces fewer crossing brackets than the vanilla inside model.

\begin{table}[t]
    \centering
    \resizebox{1.\columnwidth}{!}{%
    \begin{tabular}{lcccc}
    \toprule
    \multirow{2}{*}{\textbf{Model}} & \multicolumn{2}{c}{\textbf{KTB-40}} & \multicolumn{2}{c}{\textbf{KTB-10}} \\
    & Mean & Max & Mean & Max \\
    \midrule
    \multicolumn{3}{l}{\emph{Trivial Baselines:}}\\
    \midrule
    \multicolumn{1}{l}{Left Branching (LB)} & 29.4 & & 51.6 & \\       
    \multicolumn{1}{l}{Right Branching (RB)} & 9.8 & & 22.9 & \\   
    \midrule
    \multicolumn{3}{l}{\emph{Unsupervised Parsing approaches:}}\\
    \midrule
    \multicolumn{1}{l}{PRPN~\citep{shen2018neural}} & 27.2 & 31.8 & 30.1 & 33.6 \\
    \multicolumn{1}{l}{URNNG~\citep{kim-etal-2019-unsupervised}} & 10 & 10.2 & 22.7 & 22.7 \\
    \multicolumn{1}{l}{DIORA~\citep{drozdov-etal-2019-unsupervised}} & 24.9 & 26.0 & 42.3 & 43.3 \\
    \multicolumn{1}{l}{DIORA-all~\citep{hong-etal-2020-deep}} & 36.4 & 40.0 & 47.1 & 48.9 \\
    \multicolumn{1}{l}{Ours (using inside)} & \cellcolor{pearDark!20}33.7 & \cellcolor{pearDark!20}36.3 & \cellcolor{pearDark!20}53.8 & \cellcolor{pearDark!20}55.9 \\
    \multicolumn{1}{l}{Ours (using inside w/ self-training)} & \cellcolor{pearDark!20}37.6 & \cellcolor{pearDark!20}39.8 & \cellcolor{pearDark!20}55.5 & \cellcolor{pearDark!20}58.2 \\ 
    \multicolumn{1}{l}{Ours (using inside and outside w/ co-training)} & \cellcolor{pearDark!20}\textbf{39.2} & \cellcolor{pearDark!20}41.1 & \cellcolor{pearDark!20}\textbf{56.7} & \cellcolor{pearDark!20}59.1\\
    \midrule
    \multicolumn{1}{l}{Upper Bound} & 76.5 & & 76.6 & \\
    \bottomrule
    \end{tabular}
    }
    \caption[Results on the KTB test set]{Evalb F$_1$ on the full (F$_1$-all) and length $\leq$ 10 (F$_1$-10) sentences of the KTB test set discarding punctuation corresponding to KTB-40 and KTB-10, respectively. We take the baseline numbers of models from \citet{li-etal-2020-empirical}. See Table~\ref{tab:evalb-params} to view the hyperparameters used for \texttt{evalb}.}
    \label{tab:ktb-results}
\end{table}

\begin{table*}[t]
   \centering
    \resizebox{0.6\linewidth}{!}{%
    \begin{tabular}{l c c c c c c c}
    \toprule
    {} & {PRPN} & {ON} & URNNG & \makecell{{Compound} \\ {PCFG}} & {S-DIORA} & \makecell{{Constituency} \\ {Test}} &  \makecell{{Our Best} \\ {Parser}} \\
    \midrule
    {SBAR} & 50.0 & 51.2 & {74.8} & 56.1 & 59.2 & 66.1 & \cellcolor{pearDark!20}\textbf{81.7} \\
    {NP} & 59.2 & 64.5 & 39.5 & 74.7 & 78.0 & \textbf{79.4} & \cellcolor{pearDark!20}73.5 \\
    {VP} & 46.7 & 41.0 & 76.6 & 41.7 & \textbf{78.9} & 68.2 & \cellcolor{pearDark!20}70.4 \\
    {PP} & 57.2 & 54.4 & 55.8 & 68.8 & 67.1 & \textbf{86.2} & \cellcolor{pearDark!20}77.8 \\
    {ADJP} & 44.3 & 38.1 & 33.9 & 40.4 & 49.1 & \textbf{62.6} & \cellcolor{pearDark!20}40.9 \\
    {ADVP} & 32.8 & 31.6 & 50.4 & 52.5 & 59.9 & {63.9} & \cellcolor{pearDark!20}\textbf{70.4} \\
    \bottomrule
    \end{tabular}
    }
    \caption[Recall of constituents by label in (\%)]{{Average recall per constituent category (i.e. label recall) in (\%). The results of PRPN, ON, URNNG, and Compound PCFG are taken from \citet{kim-etal-2019-compound}, S-DIORA from \citet{drozdov-etal-2020-unsupervised}, and Constituency Test from \citet{cao-etal-2020-unsupervised}}.}
    \label{tab:label-recall}
\end{table*}

\subsection{Effect of Self-training}
\label{ssec:effect-of-selftraining}
PLMs that possess rich contextualized textual representations can assist parsing when we have a large volume of unlabeled data. For this reason, we might expect that self-training in combination with pre-training adds no extra information to the fine-tuned parser. However, we find that self-training improves the performance of the parser by about 9.8\%, demonstrating that self-training provides advantages complementary to the pre-trained contextualized embeddings (see Table~\ref{tab:self-training} in Appendix for a more detailed analysis at different stages).


\subsection{Effect of Co-training}
\label{ssec:effect-of-cotraining}
The question of how to integrate multi-view information is important. One of the options would be to concatenate both the inside and outside vectors while performing training and inference. With this experiment setting, we see negligible improvement as it only scores 13.2 F$_1$ on the PTB test (without self-training). The whole idea of separating the two models for co-training is to learn constituent boundaries to identify the splitting points in a sentence through independent views of data. This  corroborates the effectiveness of co-training compared with concatenation: the simple concatenation strategy cannot fully harvest the information corresponding to each view and indeed render the optimization intractable. After co-training, the parser achieves 63.1 F$_1$ averaged over four runs, outperforming the previous best-published result (see Table~\ref{tab:co-training} in Appendix to view the improvement at each step). Figure~\ref{fig:sent_len_vs_f1} in Appendix compares the performance of different models over varying sentence length (see Figure~\ref{fig:algo_sent_len_vs_f1} in Appendix to understand the extent to which bootstrapping helps compared to the vanilla inside model).

\subsection{Effect of Distituent Selection}
\label{ssec:effect-of-dist-select}
To understand the extent to which the type of the disitituent selection impacts the performance, we assess two settings on the \ptb{} -- random and left-branching bias. In the random setting, we select distituents from the slice \begin{small}\texttt{(start:$r$)}\end{small}, where $r$ is a random number generated between \begin{small}\texttt{start+1}\end{small} and \begin{small}\texttt{end-1}\end{small}, both inclusive. This produces 19.3 F$_{1}$ for the inside model. Whereas, in the left-branching bias setting, we prepare the seed bootstrapping process as explained in the Section~\ref{ssec:seed-bootstrap} similar to \ktb{} (a left-branching treebank). This results in 11.2 F$_{1}$ score for the inside model. Hence, the manner in which we perform the initial classification has a strong impact on the final tree structures.


\subsection{Linguistic Error Analysis}
\label{ssec:error-analysis}

Table~\ref{tab:label-recall} shows that our model achieves strong accuracy while predicting all the phrase types except for the Adjective Phrase (ADJP). We list some of the most common mistakes our parse makes and suggest likely explanations for each:

\para{Bracketing inner NP of a definite Noun Phrase.} When a definite article is linked with a singular noun, the inner spans need to be shelved, accommodating the larger span with the definite article. E.g.: \textit{the} [ \textit{stock market} ] 
\para{Grouping NP too early overlooking broader context.} Due to the way it is trained, the parser aggressively groups rare words in the corpus. Building a better outside model can fix this type of error to a considerable extent. E.g.: \textit{Shearson} [ \textit{Lehman Hutton} ] \textit{Inc.}
\para{Omitting conjunction joining two phrases.} It shows poor signs of understanding co-ordination cases in which conjunction is an adjacent sibling of the nodes being shifted, or is the leftmost or rightmost node being shifted. E.g.: \textit{Notable} [ \textit{\& Quotable} ]
\para{Confusing contractions with Possessives.} Due to the presence of a lot of contraction phrases like (\textit{they're}, \textit{it's}), the parser confuses it with that of the Possessive NPs, causing unnecessary splitting. Expanding the contractions can be a good way to correct these systematic errors. E.g.: \textit{the company} [ \textit{'s \$ 488 million in 1988} ]

In the future, we would like to develop error analysis protocols for both CTB and KTB using human-in-the-loop process (leveraging feedback from the respective language experts) and provide an in-depth statistical analysis.

\section{Related Work}
\label{sec:related-work}

Recently, neural network-based approaches have shown promising results on inducing parse trees directly from words. Our weakly-supervised parser is comparable in behavior to a fully unsupervised parser as it does not rely on syntactic annotations. We highlight some themes most relevant to our method.

\para{Learning from distant supervision.} A related work to ours~\citep{shi-etal-2021-learning} uses answer fragments and webpage hyperlinks to mine syntactic constituents for parsing. Many previous studies depend on punctuation as a strong signal to detect constituent boundaries~\citep{spitkovsky-etal-2013-breaking,parikh-etal-2014-spectral}.

\para{Incorporating bootstrapping techniques.} Co-training \citep{yarowsky-1995-unsupervised,10.1145/279943.279962,abney2007self} and self-training \citep{steedman-etal-2003-bootstrapping,mcclosky-etal-2006-effective,cohen-smith-2010-viterbi} are bootstrapping methods that attempt to convert a fully unsupervised learning problem to a semi-supervised learning form. More recently, \citet{mohananey-etal-2020-self, shi-etal-2020-role, steedman-etal-2003-bootstrapping} have shown the benefits of using self-training as a standard post-hoc processing step for unsupervised parsing models.

\para{Using inside-outside representations constructed with a latent tree chart parser.} Drawing inspiration from the inside-outside algorithm \citep{Baker:79}, DIORA \citep{drozdov-etal-2019-unsupervised} optimizes an autoencoder objective and computes a vector representation for each node in a tree by combining child representations recursively. To recover from errors and make DIORA more robust to local errors when computing the best parse in the bottom-up chart parsing, an improved variant of DIORA, S-DIORA \citep{drozdov-etal-2020-unsupervised} achieves it.

\para{Inducing tree structure by introducing an inductive bias to RNNs.} PRPN \citep{shen2018neural} introduces a neural parsing network that has the ability to make differentiable parsing decisions using structured attention mechanism to regulate skip connections in an RNN. ON-LSTM \citep{shen2018ordered} enables hidden neurons to learn information by a combination of gating mechanism as well as activation function. In URNNG,~\citet{kim-etal-2019-unsupervised} employs parameterized function over latent trees to handle intractable marginalization and inject strong inductive biases for the unsupervised learning of the recurrent neural network grammar (RNNG; \citealt{dyer-etal-2016-recurrent}). \citet{peng-etal-2019-palm} introduces PaLM that acts as an attention component on top of RNN.

\para{Enhancing PCFGs.} Compound PCFG \citep{kim-etal-2019-compound} which consists of a Variational Autoencoder (VAE) with a PCFG decoder, found the original PCFG is fully capable of inducing trees if it uses a neural parameterization. \citet{jin-etal-2019-unsupervised} show that the flow-based PCFG induction model is capable of using morphological and semantic information in context embeddings for grammar induction. \citet{zhu-etal-2020-return} proposes neural L-PCFGs to simultaneously induce both constituents and dependencies.

\para{Concerning PLMs.} Tree Transformer \citep{wang-etal-2019-tree} adds locality constraints to the Transformer encoder's self-attention such that the attention heads resemble a tree structure.~\citet{Kim2020Are} extract trees from pre-trained transformers. 


\para{Refining based on constituency tests.} With the help of transformations and RoBERTa model to make grammaticality decisions, \citep{cao-etal-2020-unsupervised} were able to achieve strong performance for unsupervised parsing. 

\section{Conclusion}
\label{sec:conclusion}
We propose a simple yet effective method of inducing constituency trees which is the first of its kind in achieving performance comparable to the supervised binary tree RNNG model and setting a new state-of-the-art result for unsupervised parsing using weak supervision. Our model generalizes to multiple languages of known treebanks. We have done comprehensive linguistic error analysis showing a step-by-step breakdown of the F$_1$ performance for the inside model, inside model with self-training, and the inside-outside model with a co-training-based approach. The effectiveness of our multi-view learning strategy is evident in our experiments. Future work could aim to augment the parser's capabilities to investigate cross-domain generalization and efficient cross-lingual transfer.

\section*{Acknowledgements}

The authors would like to thank the anonymous reviewers, Alexandra Birch, Frank Keller, Ankur Parikh, Marcio Fonseca, Ronald Cardenas, Zheng Zhao and Yftah Ziser for their feedback on earlier versions of this work.

\bibliography{anthology,custom}
\bibliographystyle{acl_natbib}

\appendix
\newpage

$\,$

\newpage

\section{Further Details}
\label{sec:appendix}

\subsection{Training Details}
\label{appx:training-details}
We use the Adam optimizer and, on the bootstrapped dataset, fine-tune \texttt{roberta-base} consisting of default 125M trainable parameters with a learning rate $3e-5$, batch size 32, maximum epochs 10, maximum sequence length 256, and gradient checkpointing for all our models. The values were chosen as default based on sequence classification tasks on the GLUE benchmark\footnote{\url{https://gluebenchmark.com/}} as mentioned in HuggingFace Transformers.\footnote{\url{https://huggingface.co/transformers/v2.3.0/examples.html\#glue}} We use a train/validation random split of 80/20 on the bootstrapped dataset which contains 100,000 sentences (50,152 for the distituent class and 49,848 for the constituent class) to monitor the validation loss and perform early stopping. The average sentence length is about 22 tokens. Note that the development set of \ptb{} is kept untouched. We set the patience value at 2. Model checkpointing, as well as logging, is carried out after every 100 steps.

We use a p3.8xlarge AWS instance with a single GPU having 64 GB memory to conduct all our experiments. The estimated training time
for the inside model is about 0.2h, inside model with self-training (3 iterations) is about 12h, and inside-outside model with co-training (2 iterations) is about 18h. While the inference time for all the models is roughly 1h.

For the Chinese monolingual experiment, we use \texttt{bert-base-chinese} which is trained on cased Chinese Simplified and Traditional text, and for Japanese monolingual experiment, we use \texttt{cl-tohoku/bert-base-japanese} which is trained on Japanese Wikipedia available at \url{https://huggingface.co/models}.

\paragraph{Training Data}
We tried several strategies to augment the \emph{distituent} class for our models, but without concrete gains. Some of those include word deletion (randomly selects tokens in the sentence and
replace them by a special token), span deletion (Same as word deletion, but puts more focus on deleting consecutive words), reordering (randomly sample several pairs of span and switch them pairwise) and substitution (sample some words and replace them with synonyms).

\subsection{Effect of Bootstrapping}
\begin{figure}[t]
   \centering
    \includegraphics[width=1.\columnwidth]{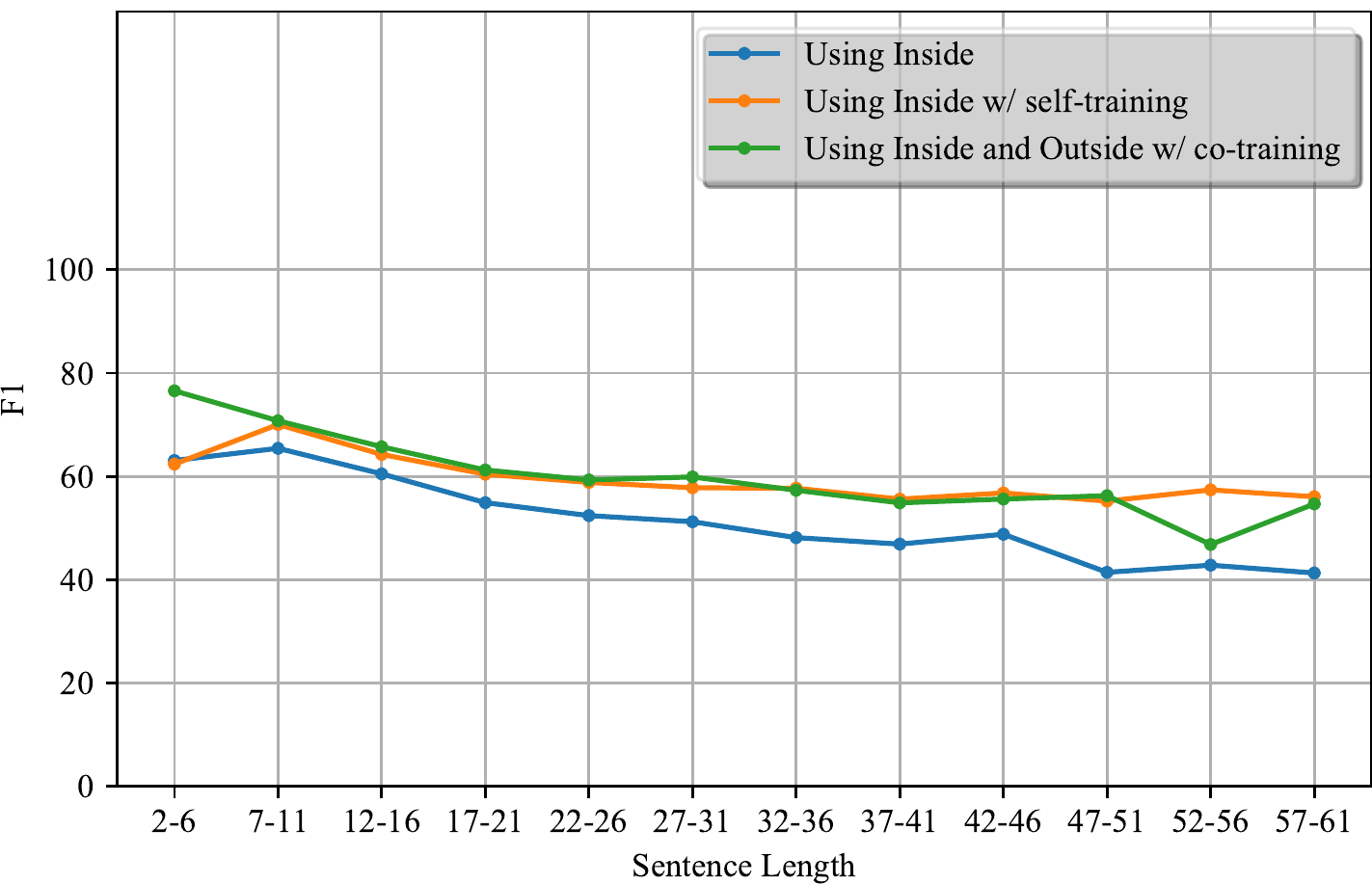}
    \caption{F$_1$ grouped by sentence length on the \ptb{} test set for different strategies.}
    \label{fig:algo_sent_len_vs_f1}
\end{figure}

As shown in Figure~\ref{fig:algo_sent_len_vs_f1}, the final model with co-training identifies constituents from shorter sentences (WSJ-10) much more precisely compared to the rest of the models. There is a lower performance in F$_1$ around sentence length of 50-55 zone, but that improves for longer sentences.\footnote{For evaluating \ptb{} and \ctb{}, we use Yoon Kim's script available at \url{https://github.com/harvardnlp/compound-pcfg}. Whereas for evaluating \ktb{}, we use Jun Li's script available at \url{https://github.com/i-lijun/UnsupConstParseEval}.}
\label{appx:bootstrapping-effect}

\subsection{Stages of Self-training}
\label{appx:stages-self-training}
Self-training boosts the performance of the inside model by 5.5 F$_1$ points as shown in Table~\ref{tab:self-training}. As can be seen, the effect of the initial set of candidate constituents and distituents on the final performance is 55.9 F$_1$ which is not insignificant.\footnote{For analysis purposes, we use the test set instead of the standard validation set to avoid tuning on the test set based on feedback received from the validation set to keep the nature of our experiments \emph{purely} unsupervised.}
\begin{table}[t]
    \centering
    \small
    \begin{tabular}{lcccc}
        \toprule
         \textbf{Model} & \multicolumn{3}{c}{ \#\textbf{{\textsc{st}}-steps}} \\
         & 0 & 1 & 2 & 3\\
        \midrule
         Inside & 55.9 & 57.7 & 59.5 & 61.4 \\
        \bottomrule
    \end{tabular}
    \caption{Unlabeled sentence-level F$_1$ on the full \ptb{} test set after applying the iterative Self-training algorithm on the Inside model.}
    \label{tab:self-training}
\end{table}

\subsection{Stages of Co-training}
\label{appx:stages-co-training}
After co-training, the performance of the inside-outside joint model increases by 1.7 F$_1$ points as shown in Table~\ref{tab:co-training}. Compared to using self-training, one of the reasons the benefit is not significant may be attributed to the fact that the inside vectors (built upon Transformer architecture) inherently possesses contextual knowledge due to being trained on a large corpus.
\begin{table}[t]
    \centering
    \small
    \begin{tabular}{lcccc}
        \toprule
         \textbf{Model} & \multicolumn{3}{c}{ \#\textbf{{\textsc{ct}}-steps}} \\
         & 0 & 1 & 2 \\
        \midrule
         \makecell{Inside and\\Outside} & 61.4 & 62.9 & 63.1 \\
        \bottomrule
    \end{tabular}
    \caption{Unlabeled sentence-level F$_1$ on the full \ptb{} test set after applying the iterative Co-training algorithm on the joint Inside and Outside model.}
    \label{tab:co-training}
\end{table}

\subsection{Unsupervised Labeled Parsing}
\label{appx:unsup-labeled-parsing}
We explore unsupervised labeled constituency parsing to identify meaningful constituent spans such as Noun Phrases (NP) and Verb Phrases (VP) to see if the parser can extract such labels. Labeled parsing is usually evaluated on whether a span has the correct label. We can effectively induce span labels using the clustering of the learned phrase vectors from the inside and outside strings. When labeling a gold bracket, our method achieves 61.2 F$_1$ on the full \ptb{} test set and is comparable with the current best model, DIORA. See Figure~\ref{fig:non-terminal-align} to view the visualization of induced and linguistic alignment.
\roberta{} does not strictly output word-level vectors. Rather, the output are subword vectors which we aggregate
with mean-pooling to achieve a word-level representation using \texttt{SentenceTransformers}.\footnote{\url{https://github.com/UKPLab/sentence-transformers}} We use 600 codes while doing the clustering initially, such that we are left with about 25 clusters after the most common label assignment process, i.e., the number of distinct phrase types. The phrase clusters are assigned to \begin{footnotesize}\textsc{\{`NP': 7, `PP': 5, `WHPP': 3, `ADVP': 3, `ADJP': 2, `S': 2, `WHADVP': 1, `UCP': 1, 'VP': 1, `PRN': 1, `QP': 1,
 `SBAR': 1, `WHNP': 1, `CONJP': 1\}}\end{footnotesize} according to the majority gold labels in that cluster. These 14 assigned phrase types correspond with the 14 most frequent labels. Table~\ref{tab:cluster-analysis}
 lists the induced non-terminal grouped across different clusters and also their correctness in identifying the gold labels. The further course of action would be to have a joint single model that is capable of achieving both bracketing and labeling. Further, these induced labels can function as features for the inside and outside models to achieve even better predictive ability. It also warrants a multi-lingual exploration in this area.

\subsection{Non-Terminal Label Alignment}
\label{appx:label-alignment}
Figure~\ref{fig:non-terminal-align} shows the alignment between gold and induced labels. We observe that some of the induced non-terminals clearly align to linguistic non-terminals. For instance, \textsc{S-2} non-terminal has a high resemblance with \textsc{NP}. Similarly, \textsc{S-8} has a high resemblance with \textsc{ADVP}. 

\begin{table}[htbp]
    \centering
    \resizebox{0.7\columnwidth}{!}{%
    \begin{tabular}{c}
    \toprule
        \multicolumn{1}{l}{\texttt{DEBUG 0}}\\
         \multicolumn{1}{l}{\texttt{MAX\_ERROR 1}}\\
         \multicolumn{1}{l}{\texttt{CUTOFF\_LEN 10}}\\
         \multicolumn{1}{l}{\texttt{LABELED 0}}\\
         \multicolumn{1}{l}{\texttt{DELETE\_LABEL\_FOR\_LENGTH -NONE-}}\\
         \multicolumn{1}{l}{\texttt{EQ\_LABEL ADVP PRT}}\\
    \bottomrule
    \end{tabular}}
    \caption[The hyperparameters used for \texttt{evalb}]{The hyperparameters used for \texttt{evalb}}.
    \label{tab:evalb-params}
\end{table}

\begin{figure}[t]
    \centering
    \includegraphics[width=1.\columnwidth]{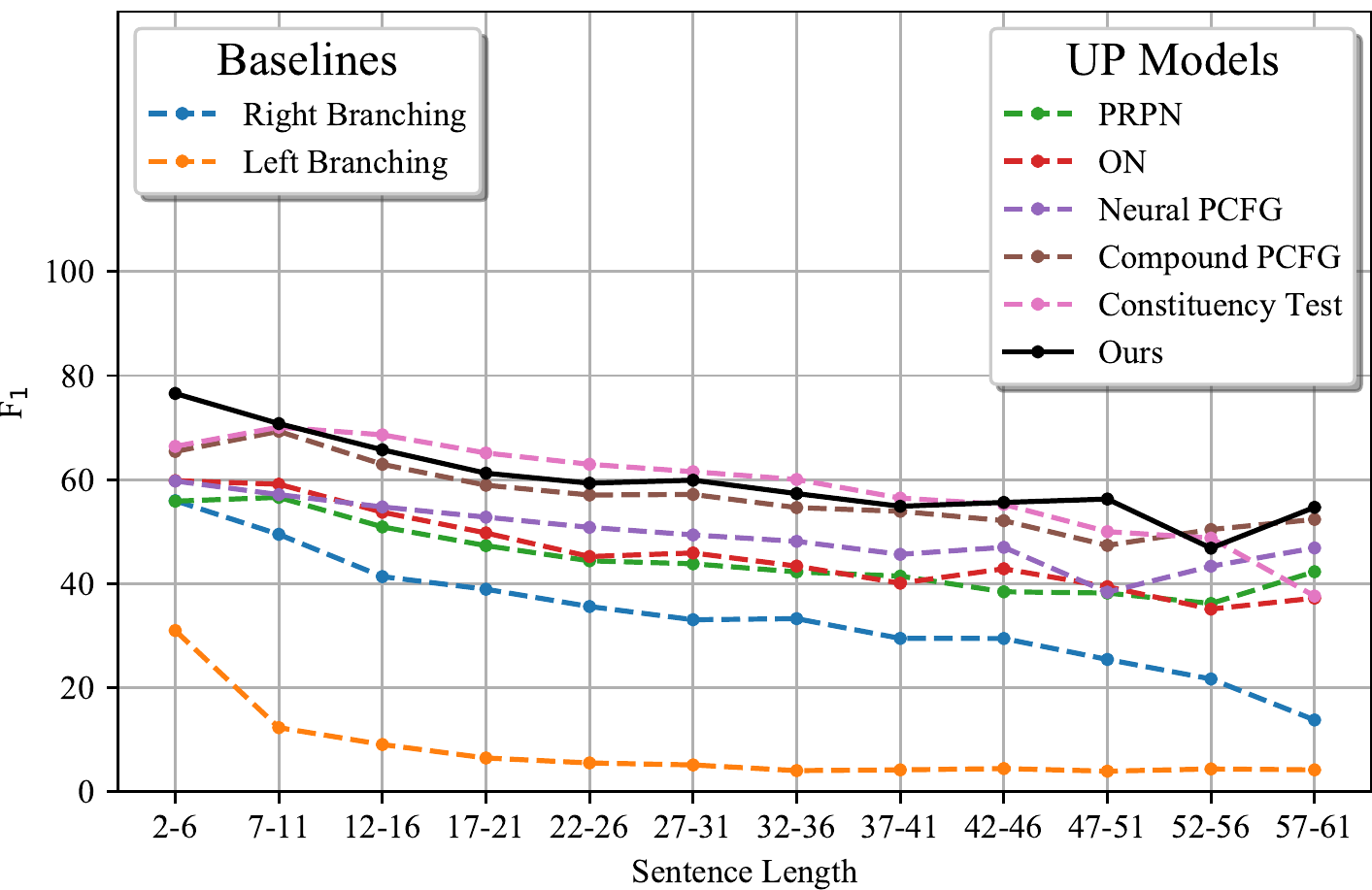}
    \caption{F$_1$ of different models grouped by sentence length on \ptb{} test set.}
    \label{fig:sent_len_vs_f1}
\end{figure}

\begin{figure*}[htb]
    \centering
    \includegraphics[height=0.9\linewidth,width=0.8\linewidth]{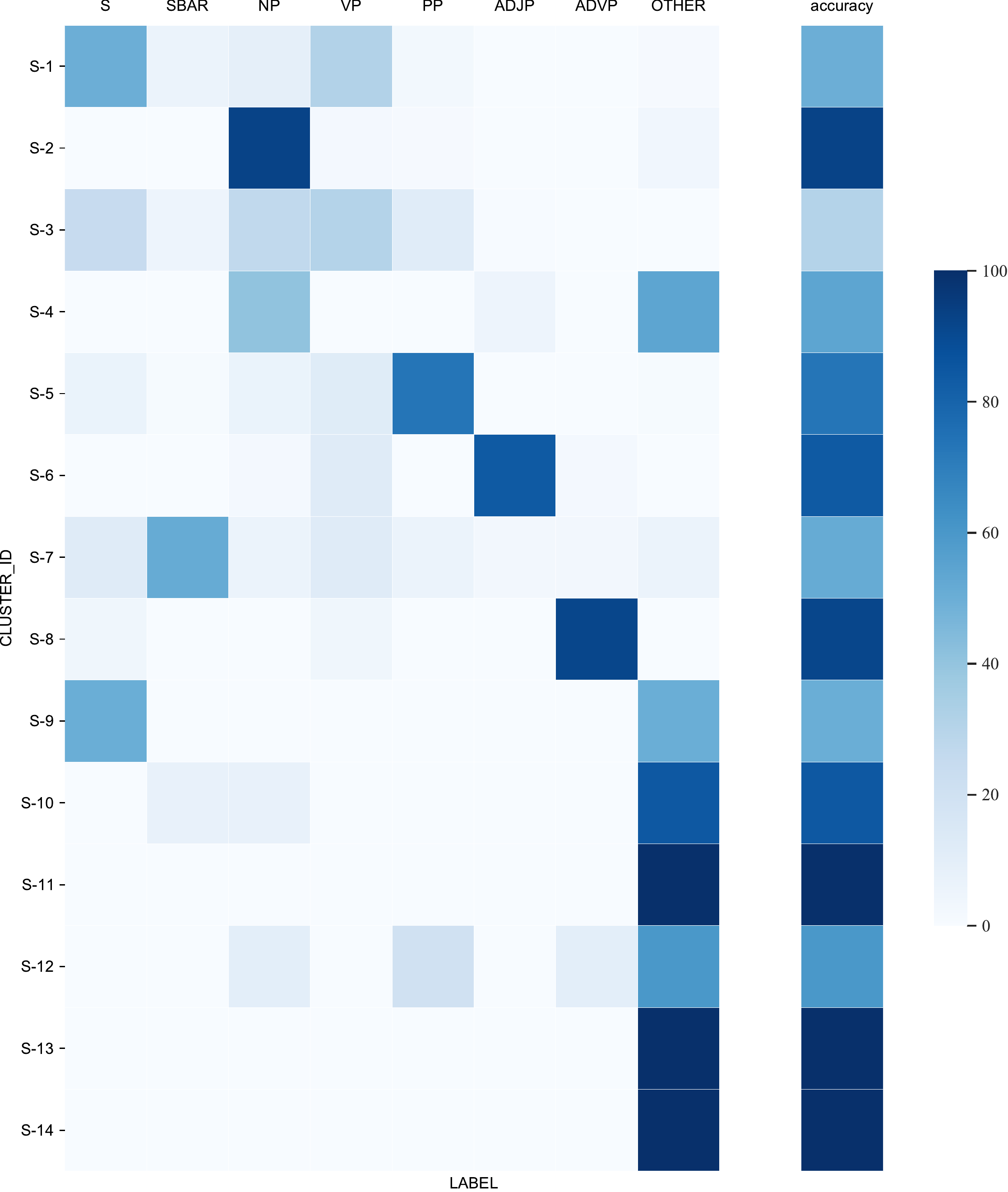}
    \caption{Alignment between induced and gold labels of the top-performing clusters. We cluster the constituent inside vectors derived from the ground truth parse (without labels) using the K-Means algorithm and assign each constituent with the most common label within its cluster. Accuracy is the probability of correctly predicting the most common label.}
    \label{fig:non-terminal-align}
\end{figure*}

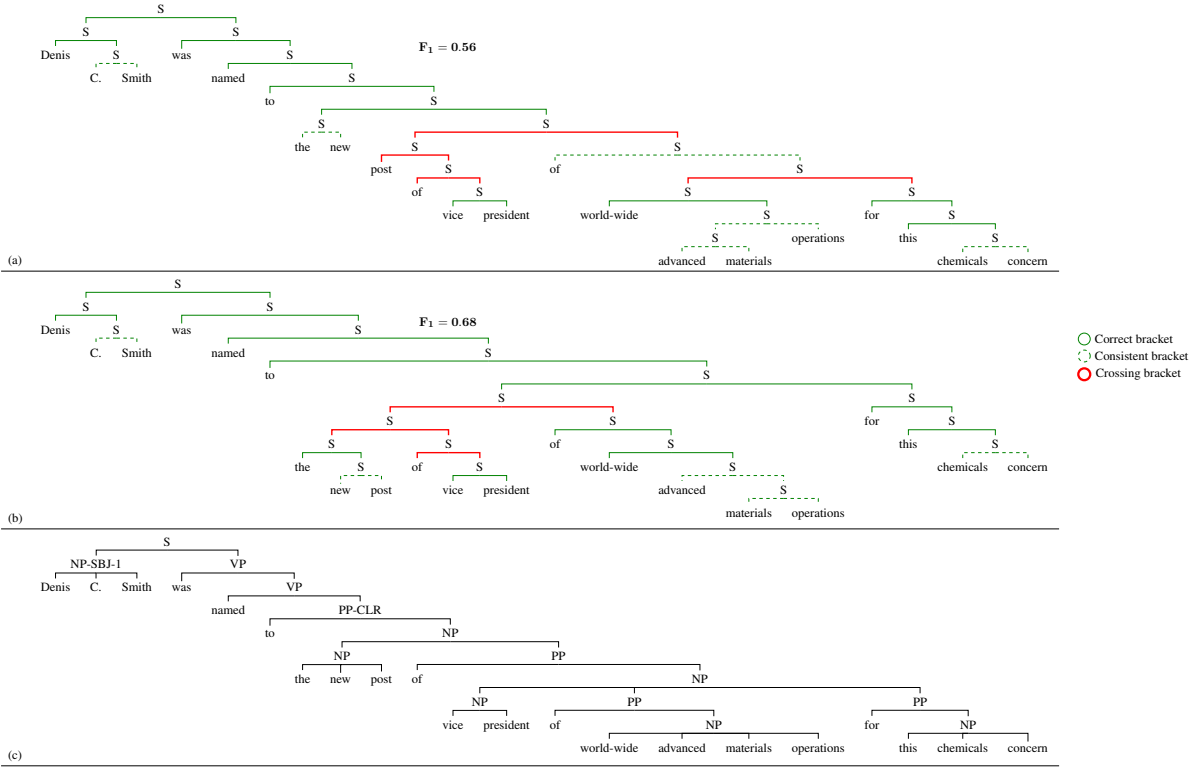
\begin{figure*}[t]
    \centering
	\resizebox{1.\linewidth}{!}{
		
		\begin{tabular}{ll}
		
		\begin{tabular}{lll}
		& \multirow{2}{*}{\begin{tikzpicture}
		\begin{scope}[xshift=3.2in,every tree node/.style={},edge from parent path={}]
            \Tree [{} {$\mathbf{F_1=0.56}$} ]]]
        \end{scope}
		\tikzset{level distance=20pt}
		\tikzset{sibling distance=10pt}
		\tikzset{edge from parent/.style=
			{draw,
				edge from parent path={(\tikzparentnode.south)
					-- +(0,0pt)
					-| (\tikzchildnode)}}}
		\Tree [ .S \edge[draw=darkgreen]; [ .S \edge[draw=darkgreen]; Denis \edge[draw=darkgreen]; [ .S \edge[dashed,draw=darkgreen]; C. \edge[dashed,draw=darkgreen]; Smith ] ] \edge[draw=darkgreen]; [ .S \edge[draw=darkgreen]; was \edge[draw=darkgreen]; [ .S \edge[draw=darkgreen]; named \edge[draw=darkgreen]; [ .S \edge[draw=darkgreen]; to \edge[draw=darkgreen]; [ .S \edge[draw=darkgreen]; [ .S \edge[dashed,draw=darkgreen]; the \edge[dashed,draw=darkgreen]; new ] \edge[draw=darkgreen]; [ .S \edge[very thick,draw=red]; [ .S \edge[very thick,draw=red]; post \edge[very thick,draw=red]; [ .S \edge[very thick,draw=red]; of \edge[very thick,draw=red]; [ .S \edge[draw=darkgreen]; vice \edge[draw=darkgreen]; president ] ] ] \edge[very thick,draw=red]; [ .S \edge[dashed,draw=darkgreen]; of \edge[dashed,draw=darkgreen]; [ .S \edge[very thick,draw=red]; [ .S \edge[draw=darkgreen]; world-wide \edge[draw=darkgreen]; [ .S \edge[dashed,draw=darkgreen]; [ .S \edge[dashed,draw=darkgreen]; advanced \edge[dashed,draw=darkgreen]; materials ] \edge[dashed,draw=darkgreen]; operations ] ] \edge[very thick,draw=red]; [ .S \edge[draw=darkgreen]; for \edge[draw=darkgreen]; [ .S \edge[draw=darkgreen]; this \edge[draw=darkgreen]; [ .S \edge[dashed,draw=darkgreen]; chemicals \edge[dashed,draw=darkgreen]; concern ] ] ] ] ] ] ] ] ] ] ]
		\end{tikzpicture}} \\[42ex]
		
        (a) & \\
		
		\midrule
		
		& \multirow{2}{*}{\begin{tikzpicture}
		\begin{scope}[xshift=3in,every tree node/.style={},edge from parent path={}]
            \Tree [{} {$\mathbf{F_1=0.68}$} ]]]
        \end{scope}
		\tikzset{level distance=20pt}
		\tikzset{sibling distance=10pt}
		\tikzset{edge from parent/.style=
			{draw,
				edge from parent path={(\tikzparentnode.south)
					-- +(0,0pt)
					-| (\tikzchildnode)}}}
		\Tree [ .S \edge[draw=darkgreen]; [ .S \edge[draw=darkgreen]; Denis \edge[draw=darkgreen]; [ .S \edge[dashed,draw=darkgreen]; C. \edge[dashed,draw=darkgreen]; Smith ] ] \edge[draw=darkgreen]; [ .S \edge[draw=darkgreen]; was \edge[draw=darkgreen]; [ .S \edge[draw=darkgreen]; named \edge[draw=darkgreen]; [ .S \edge[draw=darkgreen]; to \edge[draw=darkgreen]; [ .S \edge[draw=darkgreen]; [ .S \edge[very thick,draw=red]; [ .S \edge[very thick,draw=red]; [ .S \edge[draw=darkgreen]; the \edge[draw=darkgreen]; [ .S \edge[dashed,draw=darkgreen]; new \edge[dashed,draw=darkgreen]; post ] ] \edge[very thick,draw=red]; [ .S \edge[very thick,draw=red]; of \edge[very thick,draw=red]; [ .S \edge[draw=darkgreen]; vice \edge[draw=darkgreen]; president ] ] ] \edge[very thick,draw=red]; [ .S \edge[draw=darkgreen]; of \edge[draw=darkgreen]; [ .S \edge[draw=darkgreen]; world-wide \edge[draw=darkgreen]; [ .S \edge[dashed,draw=darkgreen]; advanced \edge[dashed,draw=darkgreen]; [ .S \edge[dashed,draw=darkgreen]; materials \edge[dashed,draw=darkgreen]; operations ] ] ] ] ] \edge[draw=darkgreen]; [ .S \edge[draw=darkgreen]; for \edge[draw=darkgreen]; [ .S \edge[draw=darkgreen]; this \edge[draw=darkgreen]; [ .S \edge[dashed,draw=darkgreen]; chemicals \edge[dashed,draw=darkgreen]; concern ] ] ] ] ] ] ] ]
		\end{tikzpicture}} \\[39ex]
	
		(b) &  \\
		
		\midrule
		
		(c) &
		
		\begin{tikzpicture}
		\tikzset{level distance=20pt}
		\tikzset{sibling distance=10pt}
		\tikzset{edge from parent/.style=
			{draw,
				edge from parent path={(\tikzparentnode.south)
					-- +(0,0pt)
					-| (\tikzchildnode)}}}
		\Tree [ .S [ .NP-SBJ-1 Denis C. Smith ] [ .VP was [ .VP named [ .PP-CLR to [ .NP [ .NP the new post ] [ .PP of [ .NP [ .NP vice president ] [ .PP of [ .NP world-wide advanced materials operations ] ] [ .PP for [ .NP this chemicals concern ] ] ] ] ] ] ] ] ]
		
		\end{tikzpicture} \\
		
		\midrule
		
		\end{tabular}
		
		& \begin{tikzpicture}[
          greendashnode/.style={shape=circle, draw=darkgreen, dashed},
          greennode/.style={shape=circle, draw=darkgreen},
          rednode/.style={shape=circle, draw=red, line width=2}
        ]
        \matrix {
              \node [greennode,label=right:Correct bracket] {}; \\
              \node [greendashnode,label=right:Consistent bracket] {}; \\
              \node [rednode,label=right:Crossing bracket] {}; \\
            };
        \end{tikzpicture}
        
        \end{tabular}
        }
		\caption[A sample parse tree taken from the PTB training set]{\label{fig:ptb-trees}Displays the parse tree output for a sample sentence: (a) Using Inside (b) Using Inside and Outside (c) Gold Tree. After the co-training procedure (b), the parser correctly identifies constituents \textit{``the new post"} and \textit{``of world-wide advanced materials operations"} which were earlier identified as distituents by the inside model (a). It makes two errors due to crossing brackets - namely \textit{``of vice president''}, \textit{``the new post of vice president"}, and \textit{``the new post of vice president of world-wide advanced materials operations".}}
\end{figure*}
\begin{CJK*}{UTF8}{gkai}
\begin{figure*}[t]
		\centering
		\resizebox{\linewidth}{!}{
		
		\begin{tabular}{lr}
		
		\begin{tabular}{lrr}
		& \multirow{2}{*}{\begin{tikzpicture}
		\begin{scope}[xshift=3in,every tree node/.style={},edge from parent path={}]
            \Tree [{} {$\mathbf{F_1=0.32}$} ]]]
        \end{scope}
		\small
		\tikzset{level distance=20pt}
		\tikzset{sibling distance=5pt}
		\tikzset{edge from parent/.style=
			{draw,
				edge from parent path={(\tikzparentnode.south)
					-- +(0,0pt)
					-| (\tikzchildnode)}}}
		\Tree [ .S \edge[draw=darkgreen]; [ .S \edge[very thick,draw=red]; [ .S \edge[very thick,draw=red]; 去年 \edge[very thick,draw=red]; 新增 ] \edge[very thick,draw=red]; [ .S \edge[very thick,draw=red]; 贷款 \edge[very thick,draw=red]; [ .S \edge[very thick,draw=red]; [ .S \edge[draw=darkgreen]; 十四点四一亿 \edge[draw=darkgreen]; 元 ] \edge[very thick,draw=red]; 比 ] ] ] \edge[draw=darkgreen]; [ .S \edge[very thick,draw=red]; 上 \edge[very thick,draw=red]; [ .S \edge[very thick,draw=red]; 年 \edge[very thick,draw=red]; [ .S \edge[draw=darkgreen]; 增加 \edge[draw=darkgreen]; [ .S \edge[draw=darkgreen]; 八亿多 \edge[draw=darkgreen]; 元 ] ] ] ] ]
		\end{tikzpicture}} \\[20ex]
		
        (a) \textit{Using Inside} & \\
		
		\midrule
		
		& \multirow{2}{*}{\begin{tikzpicture}
		\begin{scope}[xshift=3in,every tree node/.style={},edge from parent path={}]
            \Tree [{} {$\mathbf{F_1=0.63}$} ]]]
        \end{scope}
		\small
		\tikzset{level distance=20pt}
		\tikzset{sibling distance=5pt}
		\tikzset{edge from parent/.style=
			{draw,
				edge from parent path={(\tikzparentnode.south)
					-- +(0,0pt)
					-| (\tikzchildnode)}}}
		\Tree [ .S \edge[draw=darkgreen]; [ .S \edge[very thick,draw=red]; [ .S \edge[very thick,draw=red]; [ .S \edge[very thick,draw=red]; [ .S \edge[very thick,draw=red]; 去年 \edge[very thick,draw=red]; 新增 ] \edge[very thick,draw=red]; [ .S \edge[very thick,draw=red]; 贷款 \edge[very thick,draw=red]; [ .S \edge[very thick,draw=red]; [ .S \edge[draw=darkgreen]; 十四点四一亿 \edge[draw=darkgreen]; 元 ] \edge[very thick,draw=red]; 比 ] ] ] \edge[very thick,draw=red]; 上 ] \edge[very thick,draw=red]; 年 ] \edge[draw=darkgreen]; [ .S \edge[draw=darkgreen]; 增加 \edge[draw=darkgreen]; [ .S \edge[draw=darkgreen]; 八亿多 \edge[draw=darkgreen]; 元 ] ] ]
		\end{tikzpicture}} \\[25ex]
		
	
		(b) \textit{Using Inside and Outside} \\
		\textit{w/ co-training} \\ 
		
		\midrule
		
		(c) \textit{Gold} &
		
		\begin{tikzpicture}
		\small
		\tikzset{level distance=20pt}
		\tikzset{sibling distance=10pt}
		\tikzset{edge from parent/.style=
			{draw,
				edge from parent path={(\tikzparentnode.south)
					-- +(0,0pt)
					-| (\tikzchildnode)}}}
		\Tree [ .S [ .IP [ .VP 去年 [ .VP 新增 贷款 [ .QP-EXT 十四点四一亿 元 ] ] ] [ .VP [ .PP 比 [ .DP 上 年 ] ] [ .VP 增加 [ .QP-EXT 八亿多 元 ] ] ] ] ]
		
		\end{tikzpicture} \\
		
		\midrule
		
		\end{tabular}
		
		& \begin{tikzpicture}[
          greendashnode/.style={shape=circle, draw=darkgreen, dashed},
          greennode/.style={shape=circle, draw=darkgreen},
          rednode/.style={shape=circle, draw=red, line width=2}
        ]
        \matrix {
              \node [greennode,label=right:Correct bracket] {}; \\
              \node [greendashnode,label=right:Consistent bracket] {}; \\
              \node [rednode,label=right:Crossing bracket] {}; \\
            };
        \end{tikzpicture}
        
        \end{tabular}
        }
		\caption{\label{fig:ctb-trees} Example tree taken from the \ctb{} training set. After the co-training procedure (b), the parser correctly identifies constituents \textit{``十四点四一亿 元"}, \textit{``新增 贷款 十四点四一亿 元"}, and \textit{``去年 新增 贷款 十四点四一亿 元"} compared to the previous step using the inside model (a). It only makes 3 errors due to crossing brackets at \textit{``贷款 十四点四一亿 元"}, \textit{``年 增加 八亿多 元"}, and \textit{``上 年 增加 八亿多 元"}.}
\end{figure*}
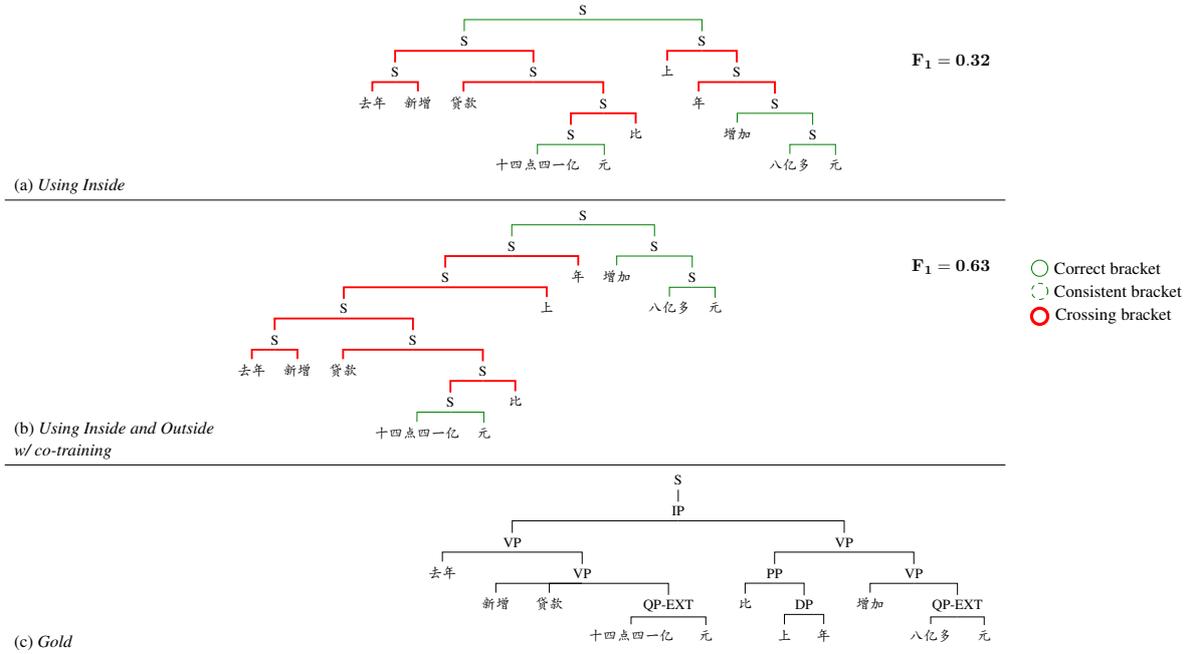
\end{CJK*}
\begin{CJK}{UTF8}{}
\begin{Japanese}
\begin{figure*}[t]
		\centering
		\resizebox{\linewidth}{!}{
		
		\begin{tabular}{lr}
		
		\begin{tabular}{lrr}
		& \multirow{2}{*}{\begin{tikzpicture}
		\begin{scope}[xshift=-6in,every tree node/.style={},edge from parent path={}]
            \Tree [{} {$\mathbf{F_1=0.13}$} ]]]
        \end{scope}
		\small
		\tikzset{level distance=20pt}
		\tikzset{sibling distance=5pt}
		\tikzset{edge from parent/.style=
			{draw,
				edge from parent path={(\tikzparentnode.south)
					-- +(0,0pt)
					-| (\tikzchildnode)}}}
		\Tree [ .S \edge[draw=darkgreen]; [ .S \edge[dashed,draw=darkgreen]; [ .S \edge[dashed,draw=darkgreen]; [ .S \edge[very thick,draw=red]; [ .S \edge[dashed,draw=darkgreen]; [ .S \edge[draw=darkgreen]; [ .S \edge[draw=darkgreen]; [ .S \edge[dashed,draw=darkgreen]; [ .S \edge[dashed,draw=darkgreen]; [ .S \edge[dashed,draw=darkgreen]; [ .S \edge[dashed,draw=darkgreen]; [ .S \edge[very thick,draw=red]; [ .S \edge[dashed,draw=darkgreen]; [ .S \edge[very thick,draw=red]; *hearer* \edge[very thick,draw=red]; そんな ] \edge[dashed,draw=darkgreen]; に ] \edge[very thick,draw=red]; 私 ] \edge[dashed,draw=darkgreen]; を ] \edge[dashed,draw=darkgreen]; *を* ] \edge[dashed,draw=darkgreen]; 信じ ] \edge[dashed,draw=darkgreen]; られ ] \edge[draw=darkgreen]; ない ] \edge[draw=darkgreen]; ならば ] \edge[dashed,draw=darkgreen]; * ] \edge[very thick,draw=red]; *pro* ] \edge[dashed,draw=darkgreen]; [ .S \edge[very thick,draw=red]; [ .S \edge[very thick,draw=red]; [ .S \edge[very thick,draw=red]; [ .S \edge[very thick,draw=red]; [ .S \edge[very thick,draw=red]; [ .S \edge[very thick,draw=red]; [ .S \edge[very thick,draw=red]; よろしい \edge[very thick,draw=red]; この ] \edge[very thick,draw=red]; 市 ] \edge[very thick,draw=red]; に ] \edge[very thick,draw=red]; セリヌンティウス ] \edge[very thick,draw=red]; [ .S \edge[very thick,draw=red]; という \edge[very thick,draw=red]; 石工 ] ] \edge[very thick,draw=red]; が ] \edge[very thick,draw=red]; *が* ] ] \edge[dashed,draw=darkgreen]; い ] \edge[draw=darkgreen]; ます ]
		\end{tikzpicture}} \\[55ex]
		
        (a) \textit{Using Inside} & \\
		
		\midrule
		
		& \multirow{2}{*}{\begin{tikzpicture}
		\begin{scope}[xshift=-5in,every tree node/.style={},edge from parent path={}]
            \Tree [{} {$\mathbf{F_1=0.45}$} ]]]
        \end{scope}
		\small
		\tikzset{level distance=20pt}
		\tikzset{sibling distance=5pt}
		\tikzset{edge from parent/.style=
			{draw,
				edge from parent path={(\tikzparentnode.south)
					-- +(0,0pt)
					-| (\tikzchildnode)}}}
		\Tree [ .S \edge[draw=darkgreen]; [ .S \edge[very thick,draw=red]; [ .S \edge[very thick,draw=red]; [ .S \edge[dashed,draw=darkgreen]; [ .S \edge[draw=darkgreen]; [ .S \edge[draw=darkgreen]; [ .S \edge[dashed,draw=darkgreen]; [ .S \edge[dashed,draw=darkgreen]; [ .S \edge[dashed,draw=darkgreen]; *hearer* \edge[dashed,draw=darkgreen]; [ .S \edge[dashed,draw=darkgreen]; [ .S \edge[dashed,draw=darkgreen]; [ .S \edge[very thick,draw=red]; そんな \edge[very thick,draw=red]; [ .S \edge[very thick,draw=red]; に \edge[very thick,draw=red]; 私 ] ] \edge[dashed,draw=darkgreen]; を ] \edge[dashed,draw=darkgreen]; *を* ] ] \edge[dashed,draw=darkgreen]; 信じ ] \edge[dashed,draw=darkgreen]; られ ] \edge[draw=darkgreen]; ない ] \edge[draw=darkgreen]; ならば ] \edge[dashed,draw=darkgreen]; * ] \edge[very thick,draw=red]; *pro* ] \edge[very thick,draw=red]; [ .S \edge[very thick,draw=red]; よろしい \edge[very thick,draw=red]; この ] ] \edge[draw=darkgreen]; [ .S \edge[very thick,draw=red]; [ .S \edge[very thick,draw=red]; [ .S \edge[very thick,draw=red]; [ .S \edge[very thick,draw=red]; [ .S \edge[very thick,draw=red]; [ .S \edge[very thick,draw=red]; 市 \edge[very thick,draw=red]; [ .S \edge[very thick,draw=red]; に \edge[very thick,draw=red]; セリヌンティウス ] ] \edge[very thick,draw=red]; [ .S \edge[very thick,draw=red]; という \edge[very thick,draw=red]; 石工 ] ] \edge[very thick,draw=red]; が ] \edge[very thick,draw=red]; *が* ] \edge[very thick,draw=red]; い ] \edge[very thick,draw=red]; ます ] ]
		\end{tikzpicture}} \\[50ex]
		
	
		(b) \textit{Using Inside and Outside} \\
		\textit{w/ co-training} \\ 
		
		\midrule
		
		(c) \textit{Gold} &
		
		\begin{tikzpicture}
		\small
		\tikzset{level distance=25pt}
		\tikzset{sibling distance=10pt}
		\tikzset{edge from parent/.style=
			{draw,
				edge from parent path={(\tikzparentnode.south)
					-- +(0,0pt)
					-| (\tikzchildnode)}}}
		\Tree [ .IP [ .PP [ .IP *hearer* [ .ADVP そんな に ] [ .PP 私 を ] *を* 信じ られ ない ] ならば ] * [ .IP *pro* よろしい ] [ .PP [ .NP この 市 ] に ] [ .PP [ .NP [ .PP セリヌンティウス という ] 石工 ] が ] *が* い ます ]
		
		\end{tikzpicture} \\
		
		\midrule
		
		\end{tabular}
		
		& \begin{tikzpicture}[
          greendashnode/.style={shape=circle, draw=darkgreen, dashed},
          greennode/.style={shape=circle, draw=darkgreen},
          rednode/.style={shape=circle, draw=red, line width=2}
        ]
        \matrix {
              \node [greennode,label=right:Correct bracket] {}; \\
              \node [greendashnode,label=right:Consistent bracket] {}; \\
              \node [rednode,label=right:Crossing bracket] {}; \\
            };
        \end{tikzpicture}
        
        \end{tabular}
        }
		\caption{\label{fig:ktb-trees}{Example tree taken from the \ktb{} training set. After the co-training procedure (b), the parser correctly identifies constituents \textit{``そんな に"}, \textit{``私 を"}, \textit{``*hearer* そんな に 私 を *を* 信じ られ ない ならば"}, \textit{``*pro* よろしい"}, \textit{``この 市"}, and \textit{``この 市 に"}}, while incorrectly tagging \textit{``セリヌンティウス という 石工 が"} as a distituent compared to the previous step using the inside model (a).}
\end{figure*}
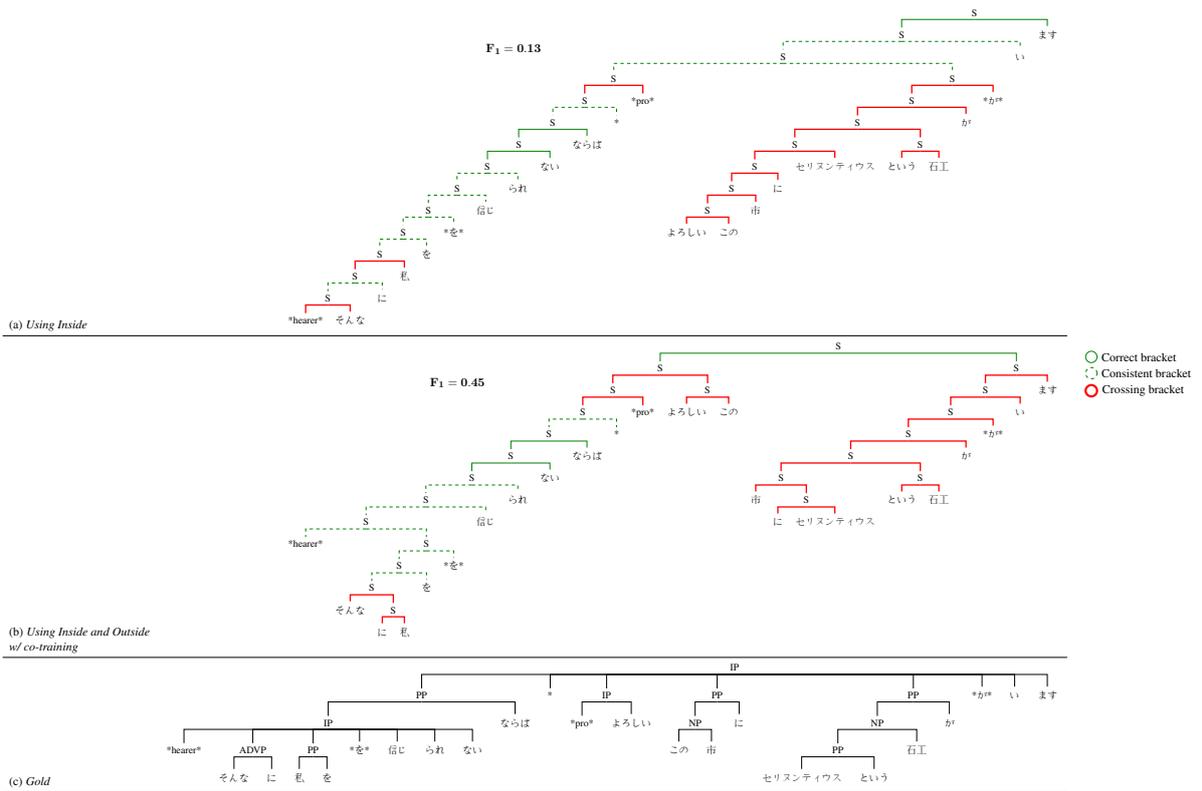
\end{Japanese}
\end{CJK}
\begin{table*}[t]
	\begin{center}
		\scalebox{0.7}{\begin{tabular}{|l|l|l|l|l|}
\hline
    &    &                                        \textbf{Constituent} & \textbf{Predicted} &  \textbf{Status} \\
\textbf{Cluster ID} & \textbf{Label} &                                                    &           &         \\
\hline
\multirow{6}{*}{0} & NP &                         the space shuttle Atlantis &        NP &    \textcolor{cadmiumgreen}{\cmark} \\
    & NP &                       Once the chief beneficiaries &        NP &    \textcolor{cadmiumgreen}{\cmark} \\
    & PP &                                      in the offing &        NP &   \textcolor{alizarin}{\xmark} \\
    & PP &                                      in the thrift &        NP &   \textcolor{alizarin}{\xmark} \\
    & S &                                the dollar was weak &        NP &   \textcolor{alizarin}{\xmark} \\
    & SBAR &                        If the new Cheer sells well &        NP &   \textcolor{alizarin}{\xmark} \\
\hline
\multirow{3}{*}{1} & ADJP &                       higher than most anticipated &        NP &   \textcolor{alizarin}{\xmark} \\
    & NP &  more than one billion Canadian dollars 851 mil... &        NP &    \textcolor{cadmiumgreen}{\cmark} \\
    & QP &                                at least 600 to 700 &        NP &   \textcolor{alizarin}{\xmark} \\
\hline
\multirow{6}{*}{12} & NP &                                    A. Boyd Simpson &        NP &    \textcolor{cadmiumgreen}{\cmark} \\
    & NP &                                Justice John Harlan &        NP &    \textcolor{cadmiumgreen}{\cmark} \\
    & NP &                                 Robert D. Cardillo &        NP &    \textcolor{cadmiumgreen}{\cmark} \\
    & NP &                                      James D. Awad &        NP &    \textcolor{cadmiumgreen}{\cmark} \\
    & NP &                              Clark S. Spalsbury Jr &        NP &    \textcolor{cadmiumgreen}{\cmark} \\
    & NP &                                        L.J. Hooker &        NP &    \textcolor{cadmiumgreen}{\cmark} \\
\hline
\multirow{3}{*}{30} & NP &                                   one 's testimony &        NP &    \textcolor{cadmiumgreen}{\cmark} \\
    & NP &                  the stock market 's plunge Friday &        NP &    \textcolor{cadmiumgreen}{\cmark} \\
    & PP &                           in the market 's decline &        NP &   \textcolor{alizarin}{\xmark} \\
\hline
\multirow{4}{*}{75} & ADVP &                                      two years ago &      ADVP &    \textcolor{cadmiumgreen}{\cmark} \\
    & ADVP &                                      two weeks ago &      ADVP &    \textcolor{cadmiumgreen}{\cmark} \\
    & PP &                            just like two years ago &      ADVP &   \textcolor{alizarin}{\xmark} \\
    & PP &                      between now and two years ago &      ADVP &   \textcolor{alizarin}{\xmark} \\
\hline
\multirow{13}{*}{310} & NP &                            action on capital gains &        VP &   \textcolor{alizarin}{\xmark} \\
    & NP &                   the three airlines being dropped &        VP &   \textcolor{alizarin}{\xmark} \\
    & NP &         news footage of the devastated South Bronx &        VP &   \textcolor{alizarin}{\xmark} \\
    & NP &      the prospect of a fight with GEC for Ferranti &        VP &   \textcolor{alizarin}{\xmark} \\
    & PP &     before declining again trapping more investors &        VP &   \textcolor{alizarin}{\xmark} \\
    & S &            This small Dallas suburb 's got trouble &        VP &   \textcolor{alizarin}{\xmark} \\
    & S &                      the earnings picture confuses &        VP &   \textcolor{alizarin}{\xmark} \\
    & SBAR &        it acquired 5 \% of the shares in Jaguar PLC &        VP &   \textcolor{alizarin}{\xmark} \\
    & SBAR &    the market is going through another October '87 &        VP &   \textcolor{alizarin}{\xmark} \\
    & VP &                         may be dubbed Eurodynamics &        VP &    \textcolor{cadmiumgreen}{\cmark} \\
    & VP &  resuscitate the protagonist of his 1972 work A... &        VP &    \textcolor{cadmiumgreen}{\cmark} \\
    & VP &                          said after the 1987 crash &        VP &    \textcolor{cadmiumgreen}{\cmark} \\
    & VP &                      has a base of 100 set in 1983 &        VP &    \textcolor{cadmiumgreen}{\cmark} \\
\hline
\multirow{10}{*}{514} & NP &                 its two classes of preferred stock &        PP &   \textcolor{alizarin}{\xmark} \\
    & NP &                             Oil company refineries &        PP &   \textcolor{alizarin}{\xmark} \\
    & PP &                         to depository institutions &        PP &    \textcolor{cadmiumgreen}{\cmark} \\
    & PP &                       of Remic mortgage securities &        PP &    \textcolor{cadmiumgreen}{\cmark} \\
    & PP &                       of the preferred-share issue &        PP &    \textcolor{cadmiumgreen}{\cmark} \\
    & PP &             in the patent-infringement proceedings &        PP &    \textcolor{cadmiumgreen}{\cmark} \\
    & PP &                             of mainframe computers &        PP &    \textcolor{cadmiumgreen}{\cmark} \\
    & PP &  from mature conventional fields in western Canada &        PP &    \textcolor{cadmiumgreen}{\cmark} \\
    & PP &             of its North American vehicle capacity &        PP &    \textcolor{cadmiumgreen}{\cmark} \\
    & VP &             have big commodity-chemical operations &        PP &   \textcolor{alizarin}{\xmark} \\
\hline
\multirow{5}{*}{533} & NP &                      Bateman Eichler Hill Richards &        NP &    \textcolor{cadmiumgreen}{\cmark} \\
    & NP &                           KLM Royal Dutch Airlines &        NP &    \textcolor{cadmiumgreen}{\cmark} \\
    & NP &                      owners Anna and Morris Snezak &        NP &    \textcolor{cadmiumgreen}{\cmark} \\
    & NP &                                      Mehta \& Isaly &        NP &    \textcolor{cadmiumgreen}{\cmark} \\
    & PP &              at Hambrecht \& Quist in San Francisco &        NP &   \textcolor{alizarin}{\xmark} \\
\hline
\end{tabular}}
\end{center}
\caption{Investigation of phrase clusters that shows several syntactic properties. Clearly, there are patterns surrounding identification of people/organization names, time-related signals, quantities etc.}
\label{tab:cluster-analysis}
\end{table*}

\ignore{
\subsection{Analyze Error Counts}
\label{appx:analyze-error}
\begin{table}[h]
    \resizebox{\columnwidth}{!}{%
    \begin{tabular}{cccccccc}
    \toprule
    {Model} & {Mod.} & {NP-I} & {NP-A} & {PP-A} & {VP-A} & {Clause} & {Coord.}\\
    \midrule
    \multicolumn{1}{l}{DIORA} & 634 & 784 & 237 & 1356 & 47 & 928 & 165 \\
    \multicolumn{1}{l}{C-PCFG} & 655 & 753 & 253 & 1997 & 42 & 858 & 166 \\
    \multicolumn{1}{l}{S-DIORA} & 487 & 917 & 265 & 861 & 91 & 954 & 186 \\
    \multicolumn{1}{l}{Ours} & 479 & & & &  \\
    \bottomrule
    \end{tabular}
    }
    \caption{Displays error counts on the \ptb{} validation set. The results of previous models are taken from \citet{drozdov-etal-2020-unsupervised}.}
    \label{tab:parse-error-analyze}
\end{table}
}

\end{document}